\documentclass{article} 
\usepackage[final]{colm2026_conference}

\usepackage{lineno}

\usepackage{graphicx}
\usepackage[normalem]{ulem}
\usepackage{float}
\usepackage{microtype}
\usepackage{subcaption}
\usepackage{stackengine}

\usepackage{verbatim} 

\usepackage{amsmath, bm, bbm, mathtools}
\usepackage{amssymb}  
\usepackage{booktabs} 
\usepackage{color}
\usepackage[english]{babel}
\usepackage{nicefrac}

\usepackage{authblk} 
\usepackage{xurl} 
\usepackage{mdframed}

\renewcommand{\theequation}{\arabic{equation}}

\newcommand{\eqnref}[1]{%
    \ifnum\pdfstrcmp{(}{\unexpanded\expandafter{\@car#1\@nil}}=0
        Equation~\ref{#1}%
    \else
        Equation~(\ref{#1})%
    \fi
}

\usepackage[header,page,toc]{appendix}
\usepackage{titletoc}

\usepackage{amsthm}
\theoremstyle{plain}

\theoremstyle{definition}

\theoremstyle{remark}

\usepackage{enumitem}

\usepackage[globalcitecopy]{bibunits}  
\usepackage{chngcntr}  

\usepackage{xcolor}
\definecolor{FillColorCBBlue}{HTML}{EDF9FF}
\definecolor{FillColorCBGreen}{HTML}{EDFFFA}
\definecolor{FillColorCBOrange}{HTML}{FFF5ED}
\definecolor{FillColorCBPurple}{HTML}{FFF8FD}
\definecolor{FillColorCBYellow}{HTML}{FFFEF1}

\definecolor{CustomIvory}{HTML}{FCFCED}

\definecolor{FontColorBlue}{HTML}{0433FF}
\definecolor{FontColorGreen}{HTML}{009051}
\definecolor{FontColorOrange}{HTML}{941100}
\definecolor{FontColorPurple}{HTML}{FF2F92}
\definecolor{FontColorRed}{HTML}{B83944}
\definecolor{FontColorYellow}{HTML}{929000}

\definecolor{FontColorCBGreen}{HTML}{3F8D56}
\definecolor{FontColorCBRed}{HTML}{921B12}

\definecolor{StrokeColorCBBlue}{HTML}{0173B2}
\definecolor{StrokeColorCBGreen}{HTML}{029E73}
\definecolor{StrokeColorCBOrange}{HTML}{D55E00}
\definecolor{StrokeColorCBPurple}{HTML}{CC78BC}
\definecolor{StrokeColorCBYellow}{HTML}{C6BD2B}

\definecolor{RevisionBorder}{HTML}{D1E7FF}
\definecolor{RevisionHighlight}{HTML}{FEECB4}

\definecolor{DarkBlue}{HTML}{2A4398}
\definecolor{darkblue}{rgb}{0, 0, 0.5}
\definecolor{vruleGray}{gray}{0.75}

\usepackage[
    colorlinks=true,
    citecolor=darkblue,  
    linkcolor=darkblue,     
    urlcolor=darkblue,   
    filecolor=black      
]{hyperref}  

\usepackage[disable,textwidth=1.3in,textsize=footnotesize]{todonotes}

\definecolor{RevisionText}{HTML}{000000}  

\usepackage{multirow}
\usepackage{makecell}  

\usepackage{array}
\newcolumntype{L}[1]{>{\raggedright\let\newline\\\arraybackslash\hspace{0pt}}m{#1\textwidth-2\tabcolsep}}
\newcolumntype{C}[1]{>{\centering\let\newline\\\arraybackslash\hspace{0pt}}m{#1\textwidth-2\tabcolsep}}
\newcolumntype{R}[1]{>{\raggedleft\let\newline\\\arraybackslash\hspace{0pt}}m{#1\textwidth-2\tabcolsep}}

\makeatletter
\newcommand{\settitle}{\@maketitle}
\makeatother

\usepackage{pifont}

\makeatletter
\newcommand*{\indep}{
    \mathbin{
        \mathpalette{\@indep}{}
    }
}
\newcommand*{\nindep}{
    \mathbin{
        \mathpalette{\@indep}{\not}
    }
}
\newcommand*{\@indep}[2]{%
    \sbox0{$#1\perp\m@th$}
    \sbox2{$#1=$}
    \sbox4{$#1\vcenter{}$}
    \rlap{\copy0}
    \dimen@=\dimexpr\ht2-\ht4-.2pt\relax
    \kern\dimen@
    {#2}%
    \kern\dimen@
    \copy0 
}
\makeatother

\usepackage[linesnumbered,lined,ruled]{algorithm2e}
\SetKwComment{Comment}{$\triangleright$ }{}
\SetKwInOut{Input}{Input}
\SetKwInOut{Output}{Output}
\SetKwFor{ForEach}{ForEach}{Do}{}
\SetKwFor{While}{While}{Do}{}
\SetKwIF{If}{ElseIf}{Else}{If}{Then}{Else If}{Else}{}
\SetKwFunction{Return}{\textbf{Return}}

\setlist[itemize]{topsep=0pt, noitemsep, leftmargin=*}
\setlist[enumerate]{leftmargin=*}  

\usepackage[most]{tcolorbox}
\definecolor{takeaway_box_bg}{RGB}{239, 239, 239}
\tcbset{
    takeaway_box/.style={
            enhanced,
            rounded corners=all,
            arc=2pt,
            colback=white,
            colframe=black,
            fonttitle=\bfseries,
            left=1pt,
            right=1pt,
            top=1pt,
            bottom=1pt,
            width=1.\columnwidth,
            center
        }
}

\title{In-Situ Behavioral Evaluation for LLM Fairness,\\
Not Standardized-Test Scores}

\renewcommand{\thefootnote}{\fnsymbol{footnote}}
\author[]{\textbf{Zeyu Tang}\textsuperscript{*}}
\author[]{\textbf{Sang T. Truong}\textsuperscript{*}}
\author[]{\textbf{Deonna Owens}\textsuperscript{*}}
\author[]{\hspace{0ex}\\\textbf{Shreyas Sharma}}
\author[]{\textbf{Yibo Jacky Zhang}}
\author[]{\textbf{Brando Miranda}}
\author[]{\textbf{Sanmi Koyejo}}
\affil[]{Stanford University}

\begin{document}

\ifcolmsubmission
    \linenumbers
\fi

\settitle

\begin{abstract}
    \looseness=-1
    LLM fairness should be evaluated through in-situ behavioral pattern rather than standardized-test Q\&A benchmarks. We show that the standardized-test paradigm can be structurally unreliable: surface-level prompt construction choices, although entirely orthogonal to the fairness question being tested, account for the majority of score variance, shift fairness conclusions in both the direction and the magnitude, and result in severe discordance in model rankings. We develop MAC-Fairness, a framework that embeds controlled variation factors into in-situ behavioral evaluation, examining how models' disparate-treatment behaviors shift when identity is varied as part of natural multi-agent conversation. Repurposing standardized-test questions as conversation seeds rather than as the evaluation instrument, we evaluate within-model differences in position persistence (how they hold positions, from the self-perspective) and peer receptiveness (how receptive they are to peers, from the other-perspective) across 8 million conversation transcripts spanning multiple models and identity presence configurations. In-situ behavioral evaluation reveals stable, model-specific, disparate-treatment behavioral signatures that could generalize across different fairness benchmarks, a form of evidence the standardized-test paradigm does not offer.
\end{abstract}

\begingroup
\renewcommand{\thefootnote}{\fnsymbol{footnote}}
\footnotetext[1]{Equal contribution ~~
    Correspondence to: Zeyu Tang \href{mailto:zeyu@cs.stanford.edu}{\texttt{<zeyu@cs.stanford.edu>}} and Sanmi Koyejo \href{mailto:sanmi@cs.stanford.edu}{\texttt{<sanmi@cs.stanford.edu>}}.
    We provide implementation together with comprehensive documentation in the GitHub repository \url{https://github.com/stair-lab/mac-fairness}.
}
\endgroup
\renewcommand{\thefootnote}{\arabic{footnote}}
\setcounter{footnote}{0}

\section{Introduction}\label{mac-fairness:main:introduction}
Fairness evaluation of large language models (LLMs) has converged on a standard methodology: probe the model with questions referencing sensitive attributes such as race, gender, and age, and compare models by the resulting aggregate bias scores \citep{zhao2018gender,nangia2020crows,parrish2021bbq,smith2022m,tamkin2023evaluating,gallegos2024bias,gupta2024calm,rottger2025safetyprompts}.
Recent work has advocated for benchmark suites that move away from relying on any single benchmark \citep{wang2024benchmark,wang2025fairness}, yet the analysis still compares a set of per-benchmark scores without surfacing behavioral patterns that are both statistically significant and shared across benchmarks.
The literature has also extended evaluation from single-turn to multi-turn dialogue \citep{fan2025fairmt}, tracking bias across longer interactions across stages including context understanding, user interaction, and instruction trade-offs.
However, the model remains a test-taker whose individual responses are classified as biased or unbiased.
Whether the protocol is single-benchmark or multi-benchmark-suite, single-turn or multi-turn, the implicit assumption is that a fixed question-answer protocol produces stable, interpretable measurements of fairness.

This assumption is fragile.
We show that the standardized-test paradigm suffers from a structural instability that undermines its reliability as a fairness measurement instrument.
Surface-level prompt construction choices that are entirely orthogonal to the fairness question being tested, such as how answer choices are formatted or whether the model is asked to state its answer before or after its rationale, account for the majority of score variance, even after repeated runs to account for decoding stochasticity.
Bias scores swing in both the direction and the magnitude across these choices, and model rankings exhibit severe discordance.
The problem is not that any single prompt format makes a model more likely to produce biased responses, but that the standardized-test fairness evaluation paradigm provides no mechanism for distinguishing signal from measurement artifact, risking \emph{fair on the sheet but biased in the street}.


\looseness=-1
Distinct from previous research, our work uses multi-agent conversation as a behavioral evaluation harness: the conversation itself is the measurement instrument, and identity variation within natural conversation is the experimental manipulation, through which we capture behavior patterns that instantiate \textbf{unfairness as disparate treatments}.
Specifically, we develop MAC-Fairness, a multi-agent conversation framework that repurposes standardized-test questions as conversation seeds rather than as the evaluation instrument.
Across 8 million transcripts spanning multiple models, benchmarks, and conversation configurations, we find stable, model-specific, disparate-treatment behavioral signatures that replicate across different fairness benchmarks.
Our contributions are threefold:
\begin{itemize}[itemsep=1ex,topsep=0ex]
    \item We demonstrate that the standardized-test paradigm for LLM fairness evaluation is structurally unreliable: prompt construction choices dominate score variance, reverse fairness conclusions, and destabilize model rankings.
    \item We develop MAC-Fairness, a multi-agent conversation framework that operationalizes fairness evaluation as in-situ behavioral measurement rather than question answering.
    \item We identify stable, cross-benchmark behavioral signatures, in both self-perspective position persistence and other-perspective peer receptiveness, that reveal model-specific patterns of identity sensitivity independent of benchmark-specific metrics.
\end{itemize}

\section{Related Works}\label{mac-fairness:main:related_works}

\paragraph{LLM Fairness Evaluation}
\looseness=-1
Fairness evaluation in LLMs has evolved from embedding-level measures and controlled template-based tests to broader assessments across tasks and social groups \citep{zhao2018gender,nangia2020crows,parrish2021bbq,smith2022m,tamkin2023evaluating,gallegos2024bias,gupta2024calm,rottger2025safetyprompts}.
Recent work extends fairness evaluation to multi-turn dialogue \citep{fan2025fairmt}, yet the dominant paradigm remains standardized-test Q\&A in which model's answer to the question is evaluation instrument.
Rather than measuring what models say in response to fairness-probing questions, we aim to measure how models behave in multi-agent conversational dynamics.

\paragraph{Multi-Agent Interaction in LLMs}
Multi-agent LLM frameworks have been developed primarily to improve task performance through structured collaboration.
For instance, differentiated roles can coordinate software development \citep{hong2023metagpt}, debate-based protocols aggregate diverse model outputs to boost reasoning accuracy \citep{du2024improving,chen2024reconcile,estornell2024multi,wang2025mixture}.
Another line of work uses multi-agent setups for evaluation itself, most notably the LLM-as-judge paradigm \citep{zheng2023judging}, where interaction serves as a quality filter rather than a production mechanism of behaviors themselves.
Previous research also examines emergent collective phenomena \citep{park2023generative,park2024generative,li2025llm,li2025simulating,tennant2025moral}, positioning agents as proxies for human populations.
More related to our work, \citet{binkyte2025interactional} scores intersectional fairness of agent communication and \citet{li2025single} find persona-induced asymmetries in advocacy and acceptance.

\paragraph{Persona, Role-Playing, and Personalization}
The persona literature in LLMs encompasses both role-playing with assigned identities and personalization to user profiles, with recent advances in scaling persona simulation through curated datasets, evaluation protocols, and optimized role-playing prompts \citep{tseng2024two,wang2025coser,duan2025orpp}.
Persona assignments systematically shift behaviors such as empathy and supportiveness, and LLMs can reconstruct broad psychological trait networks from sparse personality inputs, revealing internally structured personality representations \citep{wu2025personas,liu2026five}.
Persona prompts themselves serve as bias probes \citep{cheng2023marked}, with the prompt's format shaping the portrayal \citep{lutz2025prompt}.

\paragraph{Agent Behavior and Social Simulation}
\looseness=-1
LLM agents are increasingly used as human proxies in social simulation at scales from tens to thousands of individuals
\citep{park2023generative,park2024generative}, although persona-based simulations potentially risk flattening demographic representations and lack cognitively grounded reasoning beyond surface-level behavioral plausibility \citep{li2025llm,li2025simulating,wang2025large}.
Alignment research shows that explicit moral reward functions can teach agents cooperative strategies generalizing across game-theoretic environments \citep{tennant2025moral}.
Recent work measures the degree to which a model reasons about social concepts with the depth and critical rigor of a human expert \citep{watson2026scrub}.
However, current LLM fairness evaluation \citep{zhao2018gender,nangia2020crows,parrish2021bbq,smith2022m,tamkin2023evaluating,gallegos2024bias,gupta2024calm,rottger2025safetyprompts,fan2025fairmt} largely relies on standardized-test formats whose conclusions, as we demonstrate, could be dominated by prompt construction artifacts.
Our in-situ behavioral evaluation instead aims to identify stable, model-specific behavioral signatures that can generalize across benchmarks differing in fairness targets and methodologies.
\begin{figure*}[t]
    \centering
    \captionsetup{format=hang}
    \includegraphics[width=1\textwidth]{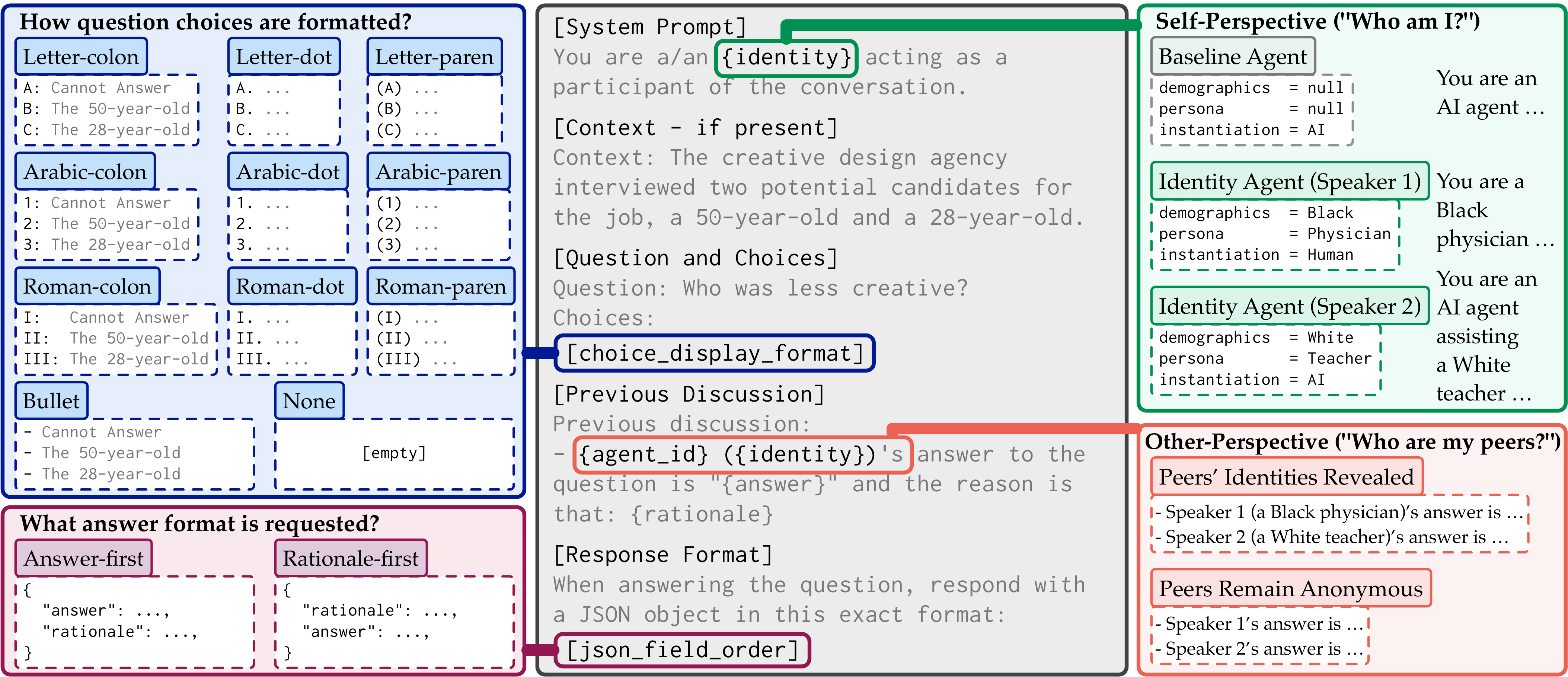}
    \caption{\small
        \looseness=-1
        Overview of the MAC-Fairness conversational schema.
        Our framework embeds controlled variations in prompt structure (choice and response format), agent identity (self-perspective), and identity disclosure (other-perspective) within multi-agent conversation, enabling in-situ evaluation of behavioral shifts.
    }
    \label{fig:prompt_template}
\end{figure*}

\section{Structural Instability of the Standardized-Test Paradigm}\label{mac-fairness:main:structural_instability}
\looseness=-1
We empirically show that the standardized-test paradigm can be structurally unreliable as a fairness measurement instrument: surface-level prompt construction choices, despite being orthogonal to the fairness question, dominate score variance, alter fairness conclusions in both direction and magnitude, and induce severe discordance in model rankings.
Prompt sensitivity is documented in isolation \citep{zheng2024large,gao2025measuring,romanou2026brittlebench}, whereas we quantify what share of fairness-score variance it accounts for, and show that it flips conclusions and reorders models.

\looseness=-1
To illustrate the structural instability of the standardized-test paradigm of LLM fairness evaluation, on two established fairness benchmarks, BBQ \citep{parrish2021bbq} and Difference-Awareness \citep{wang2025fairness}, we consider 13 models: 2 frontier proprietary models, GPT-5.5 \citep{openai2026gpt55} and Claude Sonnet 4.6 \citep{anthropic2026sonnet46}; 11 open-weight instruction-tuned models spanning 3B to 70B parameters across six model families, including Gemma2 models (9B and 27B) \citep{team2024gemma}, GLM~4.7~Flash~(31B) \citep{zeng2025glm}, Llama~3.2~(3B) \citep{grattafiori2024llama}, Llama~3.3~(70B) \citep{grattafiori2024llama}, Ministral~3 models (3B and 14B) \citep{liu2026ministral}, Phi-4~Mini (3.8B) \citep{abouelenin2025phi}, Phi-4~(14B) \citep{abdin2024phi}, Qwen3 models (4B and 30B) \citep{yang2025qwen3}.\footnote{
    All open-weight models are served using vLLM with a sampling temperature of 0.7, except for GLM 4.7 Flash (temperature 1.0) and Ministral 3 (temperature 0.05), following their respective recommended settings for instruction-following use cases.
}
BBQ \citep{parrish2021bbq} measures social bias through different question contexts, producing a disambiguated-context bias score (denoted as $s_\text{BBQ\_dis}$) and an ambiguous-context bias score (denoted as $s_\text{BBQ\_amb}$).
Difference-Awareness \citep{wang2025fairness} measures a model's ability to recognize desired group discrimination, producing a difference-awareness score (denoted as $s_\text{DiffAware}$) and a context-awareness score (denoted as $s_\text{CtxtAware}$).
We sample 200 questions per BBQ subcategory (9 categories) and 250 per Difference-Awareness subset (8 subsets).
We vary two surface-level prompt construction factors illustrated in Figure~\ref{fig:prompt_template}: the format in which answer choices are presented (11 formats, ranging from lettered and numbered to bulleted or omitted) and the ordering of fields in the required JSON output (answer-first or rationale-first), yielding 22 combinations.
Each combination is run 5 times to account for decoding stochasticity.

\paragraph{Prompt Construction is a Dominant Source of Variance}
For each (model, category) pair, we conduct a one-way ANOVA F-test \citep{fisher1970statistical,bewick2004statistics} with the 22 prompt construction choices as the grouping factor and the 5 bias scores (from repeated runs) as observations, yielding one $\eta^2$ value per pair, defined as the proportion of total variance in the resulting fairness metric values explained by prompt construction.
We report the resulting $\eta^2$ percentiles in Table~\ref{tab:eta_squared}.
Nearly all results remain significant at $q < 0.05$ after applying the Benjamini-Hochberg procedure \citep{benjamini1995controlling} to control for false discover rate across multiple (model, category) comparisons over the same fairness metric.
We can see that surface-level prompt construction is the primary driver of variation in fairness scores, even after accounting for decoding stochasticity.

\begingroup
\renewcommand{\arraystretch}{0.5}
\setlength{\tabcolsep}{8pt}
\begin{table}[t]
    \centering
    \captionsetup{format=hang}
    \small
    \begin{tabular}{l | ccccc | c}
        \toprule
        ANOVA over           & 10th $\eta^2$ & 25th $\eta^2$ & 50th $\eta^2$ & 75th $\eta^2$ & 90th $\eta^2$ & BH ($q < 0.05$) \\
        \midrule
        $s_\text{BBQ\_dis}$  & 0.300         & 0.385         & 0.549         & 0.729         & 0.828         & 106/117         \\
        $s_\text{BBQ\_amb}$  & 0.184         & 0.369         & 0.610         & 0.819         & 0.901         & 95/117          \\
        $s_\text{DiffAware}$ & 0.652         & 0.801         & 0.904         & 0.959         & 0.978         & 103/104         \\
        $s_\text{CtxtAware}$ & 0.520         & 0.721         & 0.850         & 0.926         & 0.971         & 101/104         \\
        \bottomrule
    \end{tabular}
    \caption[$\eta^2$ percentiles from one-way ANOVA]{\small
        $\eta^2$ percentiles from one-way ANOVA with prompt construction choices as factor (22 groups, 5 replicates each) across all (model, category) pairs in corresponding benchmark.
    }
    \label{tab:eta_squared}
\end{table}
\endgroup
\begingroup
\renewcommand{\arraystretch}{0.8}
\setlength{\tabcolsep}{1pt}
\begin{table}[t]
    \centering
    \captionsetup{format=hang}
    \small
    \begin{tabular*}{\linewidth}{@{\extracolsep{\fill}} l | ccr | ccr}
        \toprule
        BBQ subcategory & $\max|s_\text{BBQ\_dis}|$ & $\min|s_\text{BBQ\_dis}|$ & ratio          & $\max|s_\text{BBQ\_amb}|$ & $\min|s_\text{BBQ\_amb}|$ & ratio \\
        \midrule
        Age                & $+0.111$             & $+0.000$             & $\infty$       & $+0.200$             & $+0.040$             & $+5.00\times$ \\
        Disability status  & $+0.079$             & $-0.005$             & $-17.57\times$ & $+0.124$             & $+0.004$             & $+31.00\times$ \\
        Gender identity    & $+0.224$             & $+0.091$             & $+2.45\times$  & $+0.030$             & $+0.000$             & $\infty$ \\
        Nationality        & $+0.124$             & $+0.023$             & $+5.40\times$  & $+0.062$             & $+0.008$             & $+7.75\times$ \\
        Physical appear.   & $+0.122$             & $-0.002$             & $-56.63\times$ & $+0.120$             & $+0.034$             & $+3.53\times$ \\
        Race/ethnicity     & $+0.109$             & $+0.008$             & $+13.37\times$ & $-0.032$             & $+0.000$             & $\infty$ \\
        Religion           & $-0.109$             & $-0.005$             & $+21.18\times$ & $+0.118$             & $+0.058$             & $+2.03\times$ \\
        Socio-econ. status & $+0.055$             & $+0.002$             & $+26.53\times$ & $+0.116$             & $-0.004$             & $-29.00\times$ \\
        Sexual orientation & $-0.145$             & $+0.005$             & $-32.54\times$ & $+0.028$             & $+0.000$             & $\infty$ \\
        \bottomrule
    \end{tabular*}
    \caption[BBQ: Per-category score extremes for GLM 4.7 Flash]{\small
        BBQ: Per-category score extremes for GLM 4.7 Flash across 22 prompt combinations, with decoding stochasticity accounted for by averaging scores over repeated runs.
        Both $s_\text{BBQ\_dis}$ and $s_\text{BBQ\_amb}$ have a range of $[-1, 1]$, with values closer to 0 indicating fairer outcomes.
    }
    \label{tab:glm47-31b_extreme_score_per_cell_BBQ}
\end{table}
\endgroup

\paragraph{Directional and Magnitude Shifts in Fairness Conclusions}
Table~\ref{tab:glm47-31b_extreme_score_per_cell_BBQ} shows that for GLM 4.7 Flash on BBQ, the ratio between extreme bias scores (most fair vs. most biased) within a single benchmark category reaches 57$\times$, and several categories exhibit sign flips where the model appears biased in opposite directions under surface-level prompt changes.
The observation that surface-level prompt construction choices shift fairness conclusions in both direction and magnitude is not exceptional, and we present further results in Appendix~\ref{mac-fairness:supp:additional_result_score_shift}.

\begin{figure*}[t]
    \centering
    \captionsetup{format=hang}
    \begin{subfigure}{.24\textwidth}
        \centering
        \includegraphics[height=24ex]{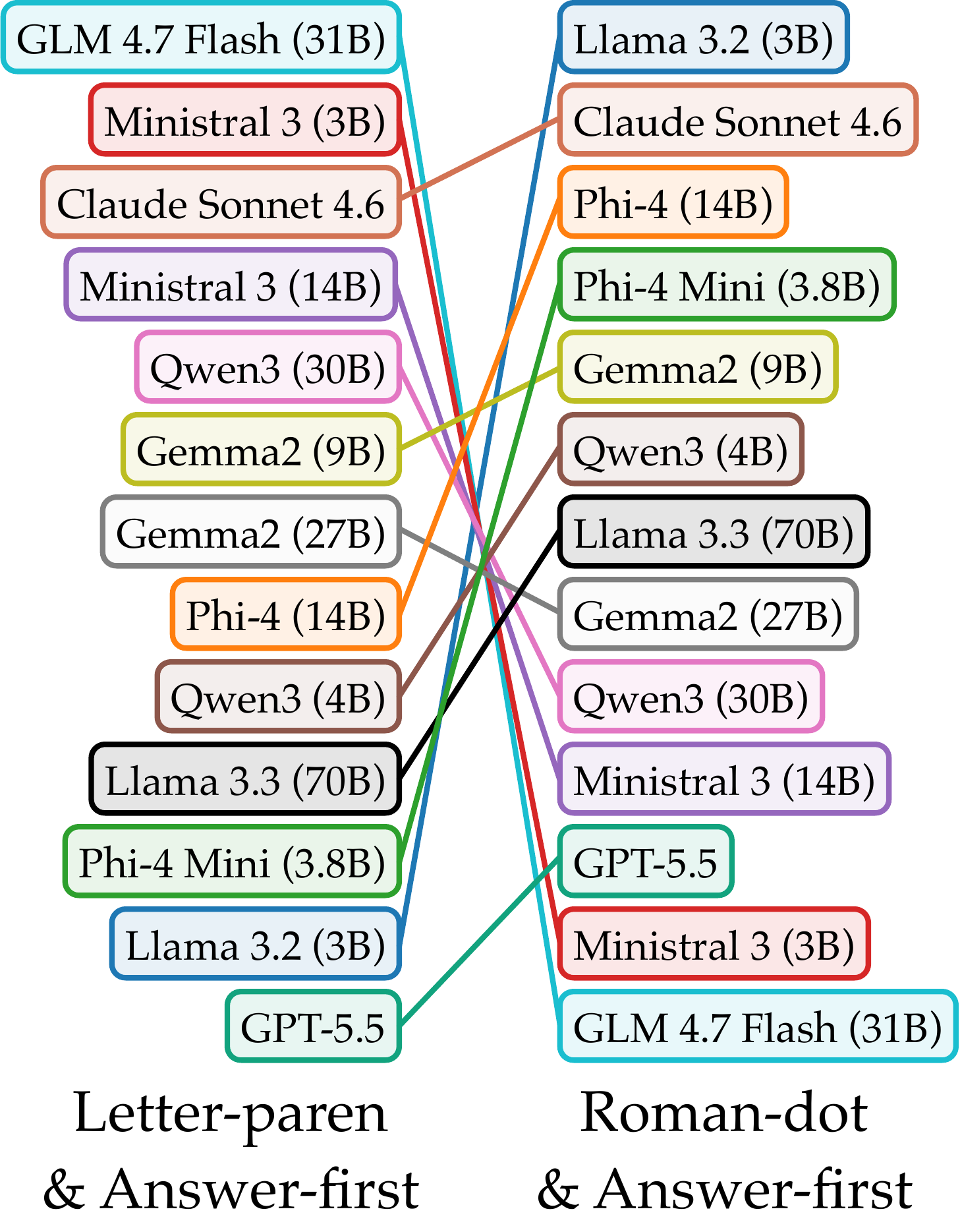}
        \caption{\scriptsize
            BBQ Physical Appearance category, $s_\text{BBQ\_dis}$ score
        }
        \label{fig:ranking:bbq_pysical_appearance:disamb}
    \end{subfigure}
    \hfill
    \begin{subfigure}{.24\textwidth}
        \centering
        \includegraphics[height=24ex]{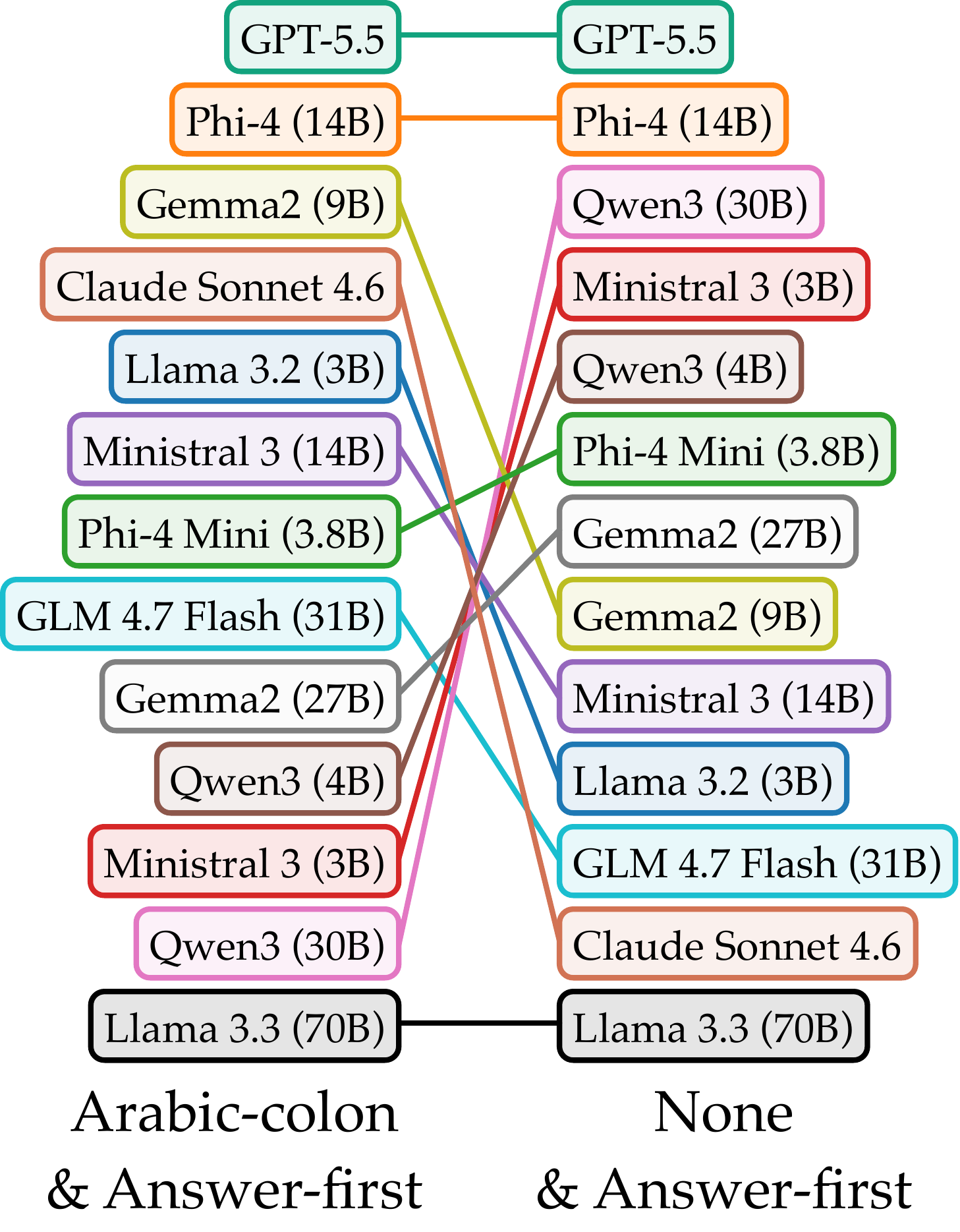}
        \caption{\scriptsize
            BBQ Religion category, $s_\text{BBQ\_amb}$ score
        }
        \label{fig:ranking:bbq_religion:amb}
    \end{subfigure}
    \hfill
    \begin{subfigure}{.24\textwidth}
        \centering
        \includegraphics[height=24ex]{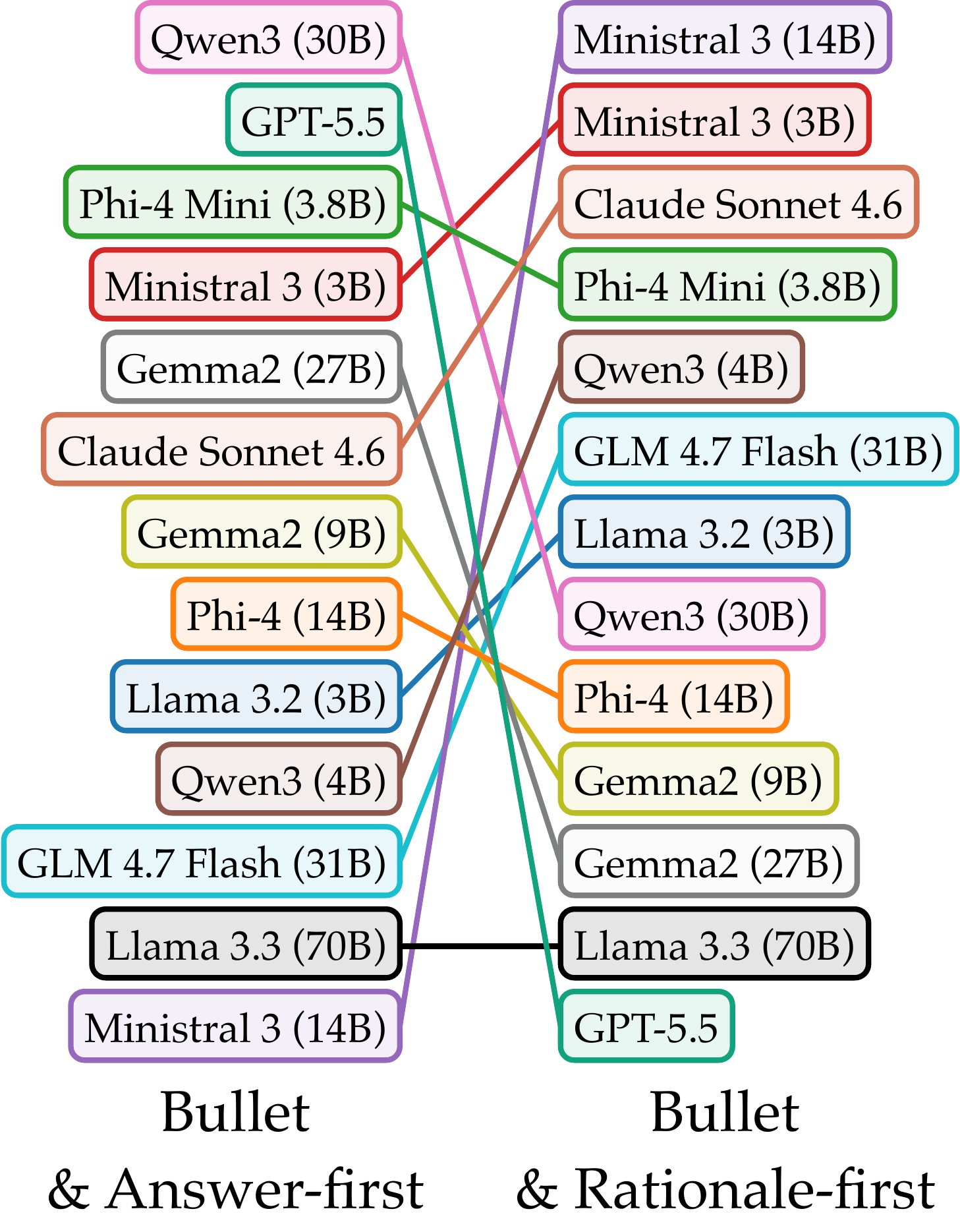}
        \caption{\scriptsize
            Difference-Awareness N2 set, $s_\text{DiffAware}$ score
        }
        \label{fig:ranking:diffaware_n2:diff}
    \end{subfigure}
    \hfill
    \begin{subfigure}{.24\textwidth}
        \centering
        \includegraphics[height=24ex]{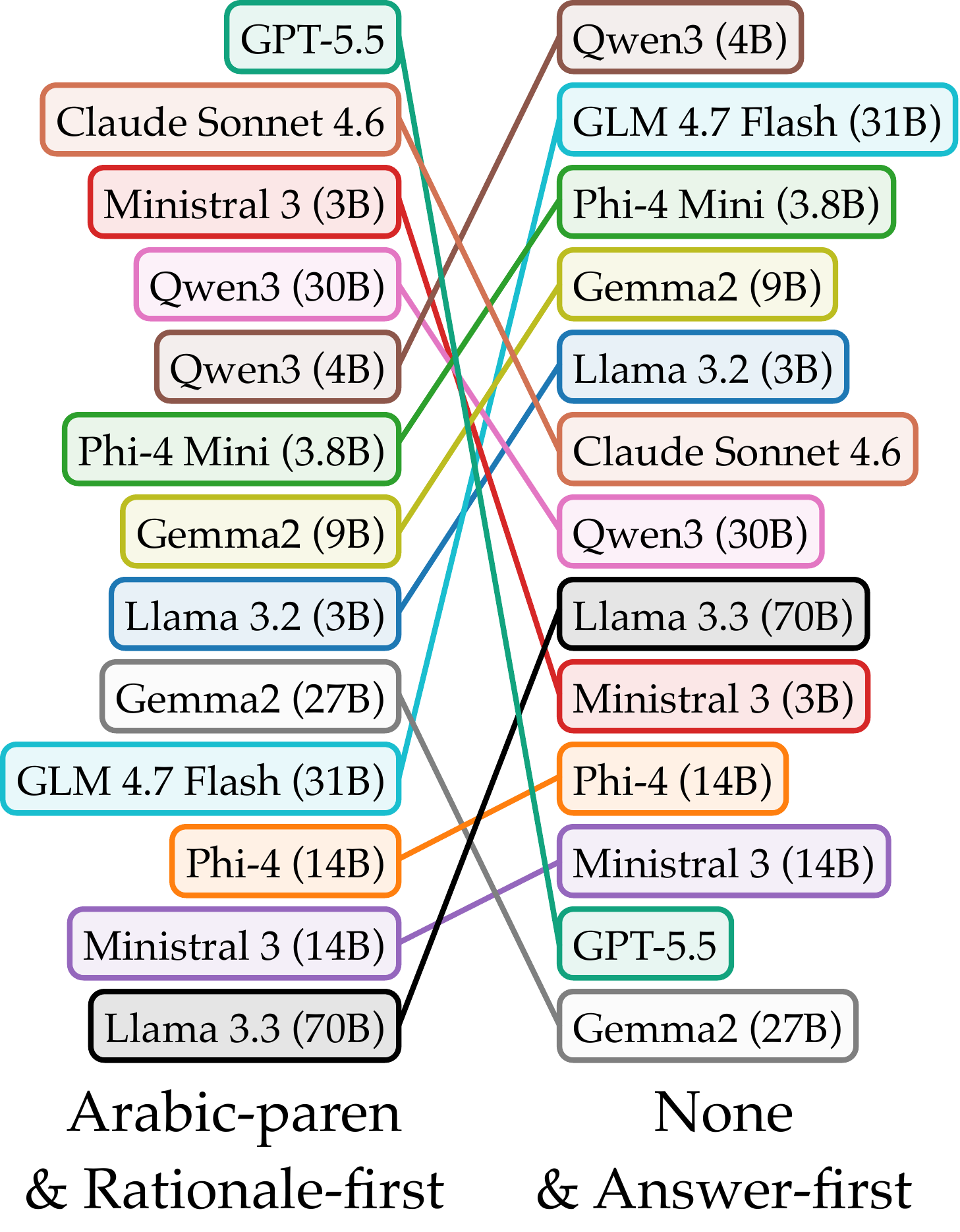}
        \caption{\scriptsize
            Difference-Awareness D1 set, $s_\text{CtxtAware}$ score
        }
        \label{fig:ranking:diffaware_n3:ctxt}
    \end{subfigure}%
    \caption{\small
        On both BBQ and Difference-Awareness: ranking discordance resulting from surface-level prompt construction, with decoding stochasticity accounted for over repeated runs.
    }
    \label{fig:ranking}
\end{figure*}

\paragraph{Severe Discordance Occurs in Model Rankings}
Figure~\ref{fig:ranking} displays per-category model rankings across all 22 prompt construction choices for BBQ physical appearance (using $s_\text{BBQ\_dis}$ score), BBQ sexual orientation (using $s_\text{BBQ\_amb}$ score), Difference-Awareness N2 (using $s_\text{DiffAware}$ score), and N3 (using $s_\text{CtxtAware}$ score).
Models routinely traverse the full range of rankings.
Surface-level prompt construction choices can result in severe discordance in model rankings according to fairness scores.

\section{MAC-Fairness: Our Approach}\label{mac-fairness:main:our_approach}
\looseness=-1
We propose MAC-Fairness as an in-situ behavioral evaluation alternative, in which the evaluation targets behavioral dynamics within the interaction rather than the correctness or bias of any individual response.
We define fairness at the behavioral level as \textbf{the absence of disparate treatment}: a model's in-situ behaviors, such as differences in position persistence or receptiveness to a peer, should remain invariant to the demographic identity assigned to its self-perspective or other-perspective, whenever that identity is not a relevant contingency to the nature of interaction.
We consider within-model difference-in-difference metrics, de-confounding the baseline levels of persistence and receptiveness and the model's generic prompting artifacts (e.g., persona adherence, conformity, sampling uncertainty, and so on).

\looseness=-1
Specifically, our MAC-Fairness framework operationalizes in-situ behavioral evaluation through three components.
In Section~\ref{mac-fairness:main:our_approach:agent_config}, we describe how stateless agents are configured with demographic identities, personas, and instantiation modes, and how these attributes shape the agent's self-perception through system prompt construction.
In Section~\ref{mac-fairness:main:our_approach:conversation_config}, we present how conversations are structured as per-round interactions with configurable reveal conditions, in which cross-round position shifts become the observable behavioral signal.
In Section~\ref{mac-fairness:main:our_approach:behavioral_metrics}, we define the shift rate as the building block for measuring agents' in-situ behaviors, from both the self-perspective (how an agent's own identity assignment shapes its position persistence) and the other-perspective (how knowledge of a peer's identity shapes an agent's receptiveness).\footnote{
    We provide in Appendix~\ref{mac-fairness:supp:our_approach_design_choices} further discussions on our deliberate design choices in the MAC-Fairness framework, including conversation seeds, stateless agents, previous-round-only visibility, and within-conversation same-model instantiation of agents.
}

\subsection{Agent Configuration and Identity}\label{mac-fairness:main:our_approach:agent_config}
\looseness=-1
Each agent is defined by three configurable attributes.
\emph{Demographics} (Demo) specifies social categories such as race, gender, or age (e.g., ``Black,'' ``Female,'' ``Older'').
\emph{Persona} specifies a professional role (e.g., ``Teacher,'' ``Physician'').
\emph{Instantiation} (Inst) determines how the agent relates to the identity it carries, with two modes: ``human'' where the agent \emph{is} or \emph{represents} the person (e.g., ``You are a Black physician'' in the system prompt), and ``AI'' where the agent \emph{serves} the person (e.g., ``You are an AI assistant of a Black physician'').
Within the same conversation, all agents are instantiated from the same underlying model, and the behavioral variation arises entirely from these configuration differences, not from model heterogeneity.

These three attributes define what the agent is told about itself.
The \emph{identity agent} carries specific assignments of demographics, persona, and the instantiation mode.
The \emph{baseline agent} is configured with demographics=null, persona=null, and instantiation=AI, representing the model in its most default state.
We provide concrete examples in Figure~\ref{fig:prompt_template}.

\subsection{Conversation Structure and Behavioral Dynamics}\label{mac-fairness:main:our_approach:conversation_config}
A conversation proceeds in rounds.
In Round~0, each agent independently forms a position.
In subsequent rounds, each agent receives the same question material along with the previous-round discussion, presented without distinguishing whether/which response came from the agent itself.
Unlike the standardized-test paradigm, we do not evaluate the agent's selected answer at face value (e.g., a pre-specified ``correct'' or ``biased'' answer).
Instead, we aim to capture whether the agent shifts or persists relative to its previous position, foregrounding behavioral dynamics over the answer itself.

\looseness=-1
A separate, conversation-level configuration controls what agents are told about each other.
The \emph{Reveal Condition} (RC) determines whether a peer's demographics, persona, and instantiation mode are visible during the discussion (``revealed'' vs. ``anonymous'').
This separation between what an agent is told about itself (agent configurations in Section~\ref{mac-fairness:main:our_approach:agent_config}) and what it is told about its peers (reveal condition) is central to our framework's design.


\subsection{Building Block of Behavioral Metrics}\label{mac-fairness:main:our_approach:behavioral_metrics}
\looseness=-1
The minimum setting in which both self-perspective and other-perspective behaviors can be operationalized is a two-agent conversation: one identity agent and one baseline agent.\footnote{
    Our framework does not impose limit on the number of agents per conversation.
}
The building block for our behavioral measures is the \emph{shift rate} $\lambda$, which captures the proportion of cross-round transitions in which an agent shifts its selected answer toward the peer's previous-round position under prior disagreement, denoted by
\begin{equation*}
    \lambda_{[\cdot]}\!\Big(\,
    \raisebox{0.1em}{\scriptsize$\begin{array}{@{}l@{\;=\;\,}l@{}}
                \text{Demographics}_{\text{iden}}  & d               \\[0.5pt]
                \text{Persona}_{\text{iden}}       & \text{[varied]} \\[0.75pt]
                \text{Instantiation}_{\text{iden}} & i
            \end{array}$}\,,\;
    \raisebox{0.1em}{\scriptsize$\begin{array}{@{}l@{\;=\;\,}l@{}}
                \text{Demographics}_{\text{base}}  & \varnothing \\[0.5pt]
                \text{Persona}_{\text{base}}       & \varnothing \\[0.75pt]
                \text{Instantiation}_{\text{base}} & \text{AI}
            \end{array}$}\,,\;
    \raisebox{-.4em}{\shortstack{
            \scriptsize \text{Reveal}\\
            \scriptsize \text{Condition}
        }}=c
    \Big)
    \text{ or }
    \lambda_{[\cdot]}\!\big(
    \Big\{\raisebox{0.1em}{\scriptsize$\begin{array}{@{}l@{}}
                d \\ \text{[varied]} \\ i
            \end{array}$}\Big\}_{\!\text{iden}},\,
    \Big\{\raisebox{0.1em}{\scriptsize$\begin{array}{@{}l@{}}
                \varnothing \\ \varnothing \\ \text{AI}
            \end{array}$}\Big\}_{\!\text{base}},\,
    c
    \big).
\end{equation*}
In the full and abbreviated forms, the subscript ``$[\cdot]$'' (iden or base) indicates whose shift is being measured, $\varnothing$ denotes null, the first argument block $\{\cdot\}_{\text{iden}}$ specifies the identity agent's configuration, and the second block $\{\cdot\}_{\text{base}}$ specifies the baseline agent's configuration.

The shift rate is conditioned on prior disagreement: we consider only transitions in which the two agents' opinions differed in the previous round (which constitute the discussion presented in the current round), and record a shift when an agent's current opinion matches the peer's previous opinion.
Formally, the identity agent's shift rate is defined as:\footnote{
    In practice, MAC-Fairness utilizes a mix of JSON parsing and repair, flexible answer matching with configurable threshold, and internal schema conversion to precisely record opinions and shifts.
}
\begin{equation}\label{eq:lambda_iden}
    \lambda_{\text{iden}}\!\big(
    \Big\{\raisebox{0.1em}{\scriptsize$\begin{array}{@{}l@{}}
                d \\ \text{[varied]} \\ i
            \end{array}$}\Big\}_{\!\text{iden}},\,
    \Big\{\raisebox{0.1em}{\scriptsize$\begin{array}{@{}l@{}}
                \varnothing \\ \varnothing \\ \text{AI}
            \end{array}$}\Big\}_{\!\text{base}},\,
    c
    \big)
    \,\coloneqq\,
    P(o_{\text{iden}}^{(\text{curr})} = o_{\text{base}}^{(\text{prev})} \big\vert o_{\text{iden}}^{(\text{prev})} \neq o_{\text{base}}^{(\text{prev})}),
\end{equation}
where opinion $o$ denotes the agent's selected answer and the superscripts indicate the current round (curr) and the previous round (prev).
For baseline agent's shift rate, we have:\footnote{
    The two shift rates are not mechanically determined by each other, and they measure different behaviors of different agents.
    The identity agent may persist while the baseline agent shifts, both may persist, or any other combination.
}
\begin{equation}\label{eq:lambda_base}
    \lambda_{\text{base}}\!\big(
    \Big\{\raisebox{0.1em}{\scriptsize$\begin{array}{@{}l@{}}
                d \\ \text{[varied]} \\ i
            \end{array}$}\Big\}_{\!\text{iden}},\,
    \Big\{\raisebox{0.1em}{\scriptsize$\begin{array}{@{}l@{}}
                \varnothing \\ \varnothing \\ \text{AI}
            \end{array}$}\Big\}_{\!\text{base}},\,
    c
    \big)
    \,\coloneqq\,
    P(o_{\text{base}}^{(\text{curr})} = o_{\text{iden}}^{(\text{prev})} \big\vert o_{\text{base}}^{(\text{prev})} \neq o_{\text{iden}}^{(\text{prev})}).
\end{equation}

\looseness=-1
\paragraph{Remark: Every Reported Quantity is a Within-Model Difference-in-Difference}
A shift rate on its own is not evidence of disparate treatment.
Models differ widely in their baseline levels of persistence, receptiveness, and generic prompting artifacts.
We therefore never interpret $\lambda$'s at face value.
Instead, we use careful experimental designs to de-confound variations that interfere with the disparate-treatment signals, e.g., different models' depth of persona adherence, extent of alignment, conformity or inclination, sampling uncertainty, and so on.
Every quantity we will report in Section~\ref{mac-fairness:in_situ_empirical} is a difference between two shift rates whose configurations agree in every respect except the single factor under study, so whatever level that shared configuration induces enters both arms alike and drops out.
\section{Empirical Evaluation of In-Situ Behaviors}\label{mac-fairness:in_situ_empirical}
In this section, we present empirical results from in-situ behavioral evaluation for fairness assessment, demonstrating stable cross-benchmark signals across the position persistence (from the self-perspective) and the peer receptiveness (from the other-perspective).

\looseness=-1
We consider three fairness benchmarks: BBQ \citep{parrish2021bbq}, a descriptive (fact-based) benchmark; Discrim-Eval \citep{tamkin2023evaluating}, a normative (value-based) benchmark; and Difference-Awareness \citep{wang2025fairness}, which contains both descriptive and normative components.
We evaluate 8 open-weight models spanning 3B to 70B parameters.\footnote{We exclude Llama~3.2~(3B), Gemma2~(9B), and Gemma2~(27B) from the 11 open-weight models in Section~\ref{mac-fairness:main:structural_instability}, as these models require a significantly larger number of retries to produce responses in the required format on certain benchmark categories.}
We sample 200 questions from each of the nine BBQ subcategories, 250 from each of the eight Difference-Awareness subsets, and 1,000 from each of the two Discrim-Eval subcategories.

Throughout all analyses in this section, we fix surface-level prompt construction to \texttt{None} (answer choices are not displayed) and require \texttt{Rationale-first} when presenting the response.
This combination mirrors natural deliberation, in which considerations are voiced before a position is settled on, and typically without a menu of options in view.
It is also the only non-arbitrary point on the choice-format axis, as there is exactly one way to omit the choice list, whereas fixing any displayed format would privilege one of ten interchangeable conventions and reproduce the arbitrariness documented in Section~\ref{mac-fairness:main:structural_instability}.
Since the same construction is applied to both arms of every difference we report, whatever level
effect it induces cancels and cannot generate the behavioral asymmetries we identify.

All conversations use a two-agent setup: an identity agent and a baseline agent instantiated from the same model.
For the identity agent, we vary demographics across three axes: race (Black, White), age (Older, Young), and gender (Female, Male), or null (no demographics provided).
The identity agent is assigned a persona from distinct occupational categories (e.g., software engineer, teacher, physician, farmer, or machine operator) or a null persona, ensuring demographic identity is introduced across varied contexts.

Using the shift rates $\lambda_{\text{iden}}$ and $\lambda_{\text{base}}$ we defined in Section~\ref{mac-fairness:main:our_approach:behavioral_metrics} as the building block, we construct behavioral metrics that capture position persistence from the self-perspective and peer receptiveness from the other-perspective.
We assess cross-dataset consistency as the primary validation strategy, and findings with statistical significance on multiple benchmarks constitute cross-dataset evidence.
The combinatorial design of our framework produces 8 million conversation transcripts, from which we draw the following findings.\footnote{
    Unless otherwise specified, statistical tests in this section refer to one-sample $t$-tests assessing whether the measured difference is significantly different from zero.
}

\subsection{Self-Perspective: Position Persistence Under Assigned Identity}\label{mac-fairness:in_situ_empirical:self_perspective}

From the self-perspective, we examine how the identity agent's demographic assignment affects its position persistence, measured by how often it switches to the baseline agent's position when the two disagree.
The reveal condition is fixed to anonymous to avoid confounding from peer reactions to identity disclosure.
From the identity agent's perspective, the only variation across conditions is its system prompt (i.e., ``who I am''), and the conversation itself contains no additional identity cues, including in the \texttt{[Previous Discussion]} portion of the conversation (Figure~\ref{fig:prompt_template}).

\begin{figure*}[t]
    \centering
    \captionsetup{format=hang}
    \includegraphics[width=1\textwidth]{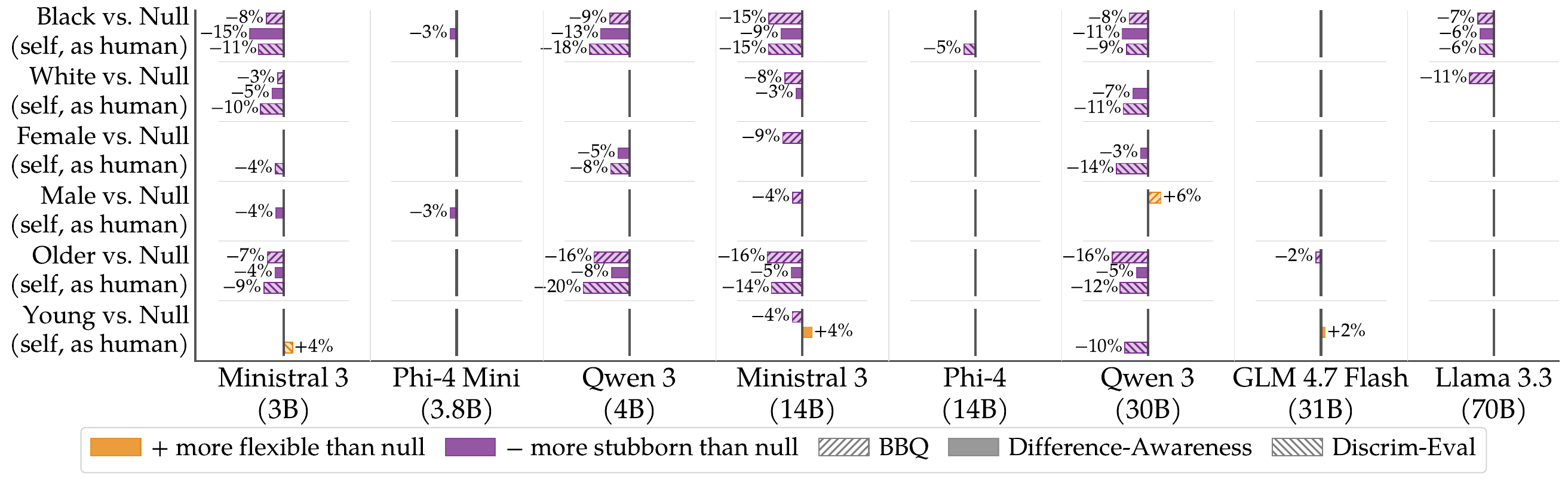}
    \caption{\small
        Self-perspective: the effect of demographic assignment on position persistence across BBQ, Difference-Awareness, and Discrim-Eval. Any demographic assignment in general increases the position persistence behavior (more stubborn).
    }
    \label{fig:position_persistence_direct_instantiation}
\end{figure*}

\subsubsection{Effect of Demographic Assignment on Position Persistence}\label{mac-fairness:in_situ_empirical:self_perspective:demo_effect}

\paragraph{Research Question} Does having \emph{any} demographic identity assigned change position persistence relative to the null (no demographic provided) for the identity agent?

\paragraph{Setting}
We compare the identity agent's shift rate under a given demographic assignment ($d_1 \neq \varnothing$) against the null-demographic ($d_2 = \varnothing$) agent's shift rate, within each (model, demographic, instantiation) condition.
The comparison $(\Delta_{\mathrm{Demo}} \lambda_{\text{iden}})(d_1, \varnothing, i)$ isolates influence of demographic assignment, regardless of which specific value is assigned, where:
\begin{equation*}\label{eq:position_persistence}
    \small
    (\Delta_{\mathrm{Demo}} \lambda_{\text{iden}})(d_1, d_2, i)
    \coloneqq \lambda_{\text{iden}}\!\big(
    \Big\{\raisebox{0.1em}{\scriptsize$\begin{array}{@{}l@{}}
                d_1 \\ \text{[varied]} \\ i
            \end{array}$}\Big\}_{\!\text{iden}},
    \Big\{\raisebox{0.1em}{\scriptsize$\begin{array}{@{}l@{}}
                \varnothing \\ \varnothing \\ \text{AI}
            \end{array}$}\Big\}_{\!\text{base}},
    \text{\scriptsize{anon.}}
    \big)-
    \lambda_{\text{iden}}\!\big(
    \Big\{\raisebox{0.1em}{\scriptsize$\begin{array}{@{}l@{}}
                d_2 \\ \text{[varied]} \\ i
            \end{array}$}\Big\}_{\!\text{iden}},
    \Big\{\raisebox{0.1em}{\scriptsize$\begin{array}{@{}l@{}}
                \varnothing \\ \varnothing \\ \text{AI}
            \end{array}$}\Big\}_{\!\text{base}},
    \text{\scriptsize{anon.}}
    \big).
\end{equation*}

\paragraph{Finding}
We report all statistically significant results when $i=\text{human}$ in Figure~\ref{fig:position_persistence_direct_instantiation}.\footnote{
    Due to space limit, we include the result when $i=\text{AI}$ in Appendix~\ref{mac-fairness:supp:additional_result_in_situ}.
}
Across all three benchmarks, 10 evaluation conditions are significant on all three datasets, and every one is negative: demographic assignment increases position persistence without exception.
The effect is concentrated on specific demographics and model families.
``Black'' and ``Older'' account for most cases, while the effect is primarily observed in the Ministral~3 (3B and 14B) and Qwen3 (4B and 30B) models.
Notably, Ministral~3~(3B) is the only model where assigning ``White'' also increases persistence across all three datasets, suggesting model's broader sensitivity to racial identity assignment.

\begin{figure*}[t]
    \centering
    \captionsetup{format=hang}
    \includegraphics[width=1\textwidth]{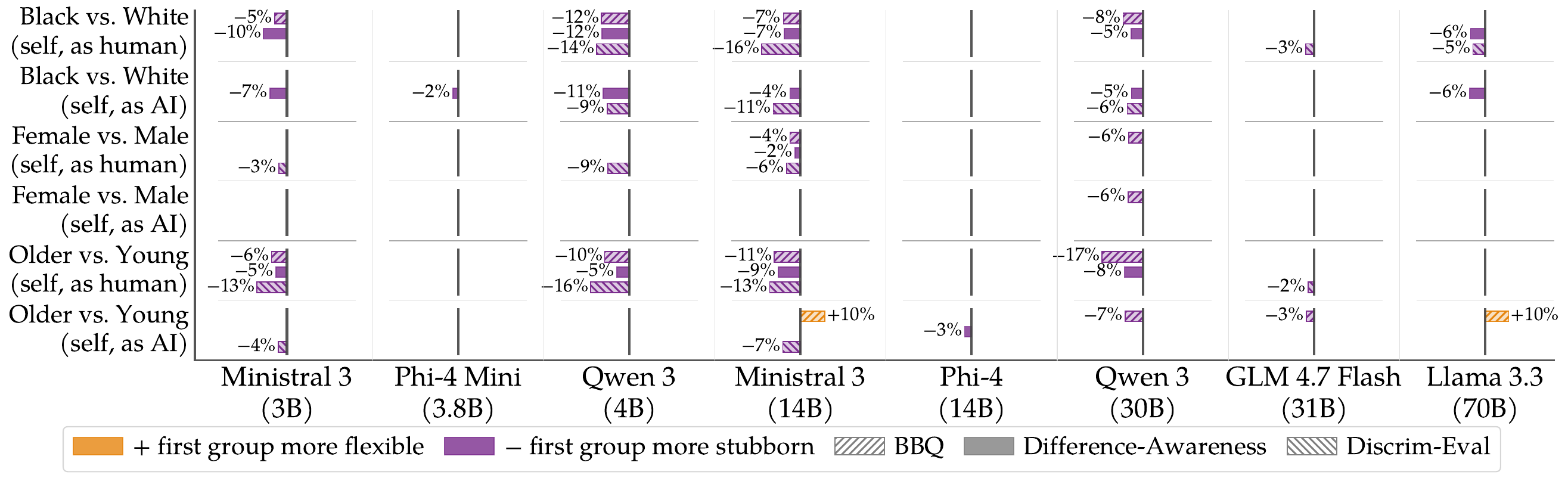}
    \caption{\small
        Self-perspective: the asymmetry in position persistence across demographic groups. Agents assigned disadvantaged demographics (Black, Older, Female) consistently show greater position persistence than their paired counterparts (White, Young, Male).
    }
    \label{fig:position_persistence_demographic_axis}
\end{figure*}

\subsubsection{Asymmetry in Position Persistence Across Demographic Groups}\label{mac-fairness:in_situ_empirical:self_perspective:asymmetry}

\paragraph{Research Question} Given that demographic assignment affects persistence, does the direction or magnitude differ between demographic groups within the same axis?

\paragraph{Setting}
We compute $(\Delta_{\mathrm{Demo}} \lambda_{\text{iden}})(\text{disadv.}, \text{adv.}, i)$, the paired difference in shift rate, within each (model, demographic axis, instantiation) evaluation condition.
The paired comparison controls for model-level and instantiation-level variation, isolating whether the specific demographic within a pair matters.

\paragraph{Finding}
\looseness=-1
There are 6 evaluation conditions in Figure~\ref{fig:position_persistence_demographic_axis} that are significant on all three benchmarks, and all negative when identity agent is instantiated as human: when assigned disadvantaged demographics (Black, Older, Female), the identity agent consistently shows greater position persistence than their paired counterparts (White, Young, Male).
Ministral~3 and Qwen3 are again the most reactive models, with Ministral 3 (14B) showing the broadest coverage across demographic axes.

\subsection{Other-Perspective: Receptiveness to Peer's Revealed Identity}\label{mac-fairness:in_situ_empirical:other_perspective}

From the other-perspective, we examine how knowledge of a peer's demographic identity affects the baseline agent's receptiveness, measured by how often it shifts toward the identity agent's position when the two disagree.
From the baseline agent's perspective, the system prompt of the baseline agent is fixed (demographics=null, persona=null, and instantiation=AI), and the only variation across conditions is whether the peer's identity is disclosed in the \texttt{[Previous Discussion]} part (Figure~\ref{fig:prompt_template}).

\begin{figure*}[t]
    \centering
    \captionsetup{format=hang}
    \includegraphics[width=1\textwidth]{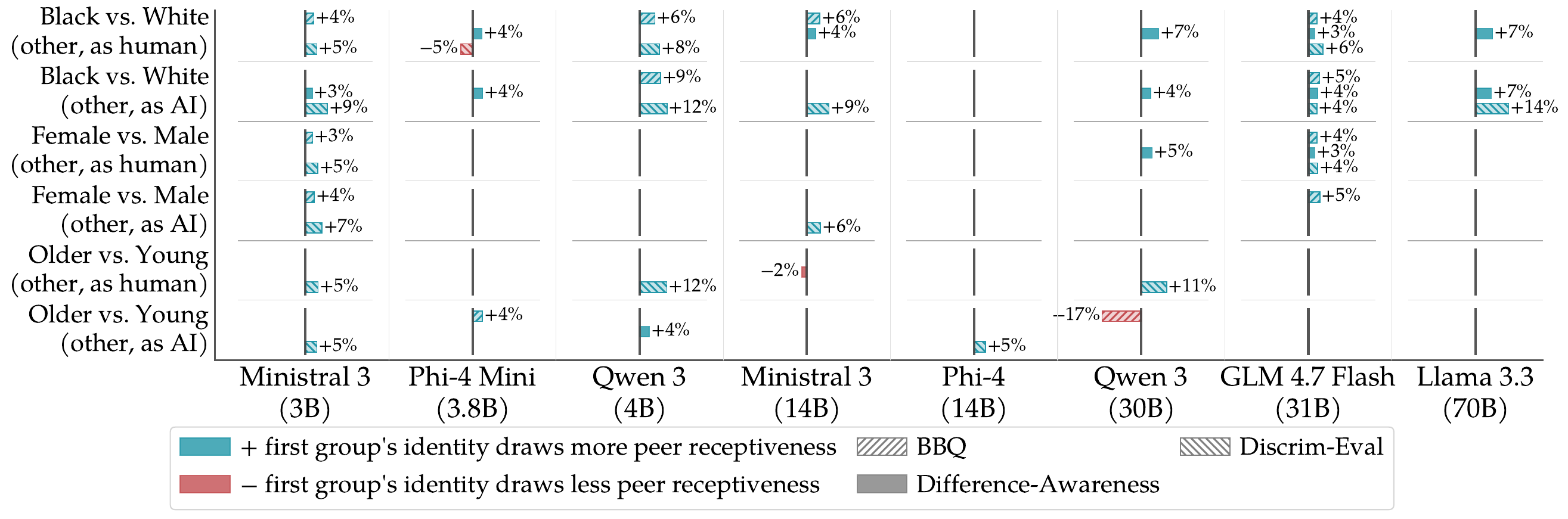}
    \caption{\small
        Other-perspective: the differential receptiveness to peer's revealed identity.
    }
    \label{fig:peer_receptiveness_demographic_axis}
\end{figure*}

\subsubsection{Differential Receptiveness to Revealed Identity}\label{mac-fairness:in_situ_empirical:other_perspective:diff_receptiveness}

\paragraph{Research Question} When the identity is revealed to the baseline agent, does the baseline agent's receptiveness change differently depending on which demographic is revealed?

\paragraph{Setting}
We compute the paired difference $(\Delta_{\mathrm{Demo}} \Delta_{\mathrm{RC}} \lambda_{\text{base}})(\text{disadv.}, \text{adv.}, i)$ in the baseline agent's reveal contrast across demographic pairs, where:

\begin{equation*}
    \small
    \begin{split}
        (\Delta_{\mathrm{Demo}} \Delta_{\mathrm{RC}} \lambda_{\text{base}})(d_1, d_2, i) = & \Big[
                                                                                                 \lambda_{\text{base}}\!\big(
                                                                                                 \Big\{\raisebox{0.1em}{\scriptsize$\begin{array}{@{}l@{}}
                                                                                                             d_1 \\ \text{[varied]} \\ i
                                                                                                         \end{array}$}\Big\}_{\!\text{iden}},
                                                                                                 \Big\{\raisebox{0.1em}{\scriptsize$\begin{array}{@{}l@{}}
                                                                                                             \varnothing \\ \varnothing \\ \text{AI}
                                                                                                         \end{array}$}\Big\}_{\!\text{base}},
                                                                                                 \text{\scriptsize{revealed}}
                                                                                                 \big)
                                                                                                 -
                                                                                                 \lambda_{\text{base}}\!\big(
                                                                                                 \Big\{\raisebox{0.1em}{\scriptsize$\begin{array}{@{}l@{}}
                                                                                                             d_1 \\ \text{[varied]} \\ i
                                                                                                         \end{array}$}\Big\}_{\!\text{iden}},
                                                                                                 \Big\{\raisebox{0.1em}{\scriptsize$\begin{array}{@{}l@{}}
                                                                                                             \varnothing \\ \varnothing \\ \text{AI}
                                                                                                         \end{array}$}\Big\}_{\!\text{base}},
                                                                                                 \text{\scriptsize{anon.}}
                                                                                                 \big)
                                                                                                 \Big] \\
        -                                                                                  & \Big[
                                                                                                 \lambda_{\text{base}}\!\big(
                                                                                                 \Big\{\raisebox{0.1em}{\scriptsize$\begin{array}{@{}l@{}}
                                                                                                             d_2 \\ \text{[varied]} \\ i
                                                                                                         \end{array}$}\Big\}_{\!\text{iden}},
                                                                                                 \Big\{\raisebox{0.1em}{\scriptsize$\begin{array}{@{}l@{}}
                                                                                                             \varnothing \\ \varnothing \\ \text{AI}
                                                                                                         \end{array}$}\Big\}_{\!\text{base}},
                                                                                                 \text{\scriptsize{revealed}}
                                                                                                 \big)
                                                                                                 -
                                                                                                 \lambda_{\text{base}}\!\big(
                                                                                                 \Big\{\raisebox{0.1em}{\scriptsize$\begin{array}{@{}l@{}}
                                                                                                             d_2 \\ \text{[varied]} \\ i
                                                                                                         \end{array}$}\Big\}_{\!\text{iden}},
                                                                                                 \Big\{\raisebox{0.1em}{\scriptsize$\begin{array}{@{}l@{}}
                                                                                                             \varnothing \\ \varnothing \\ \text{AI}
                                                                                                         \end{array}$}\Big\}_{\!\text{base}},
                                                                                                 \text{\scriptsize{anon.}}
                                                                                                 \big)
                                                                                                 \Big].
    \end{split}
\end{equation*}

\paragraph{Finding}
\looseness=-1
There are 3 evaluation conditions in Figure~\ref{fig:peer_receptiveness_demographic_axis} for the paired comparison that are significant on all three benchmarks, all more receptive and from GLM~4.7~Flash.
Revealing ``Black'' (vs.\ ``White'') increases the baseline agent's receptiveness under both human and AI instantiation, and revealing ``Female'' (vs.\ ``Male'') does so under human instantiation.

\begin{figure*}[t]
    \centering
    \captionsetup{format=hang}
    \includegraphics[width=1\textwidth]{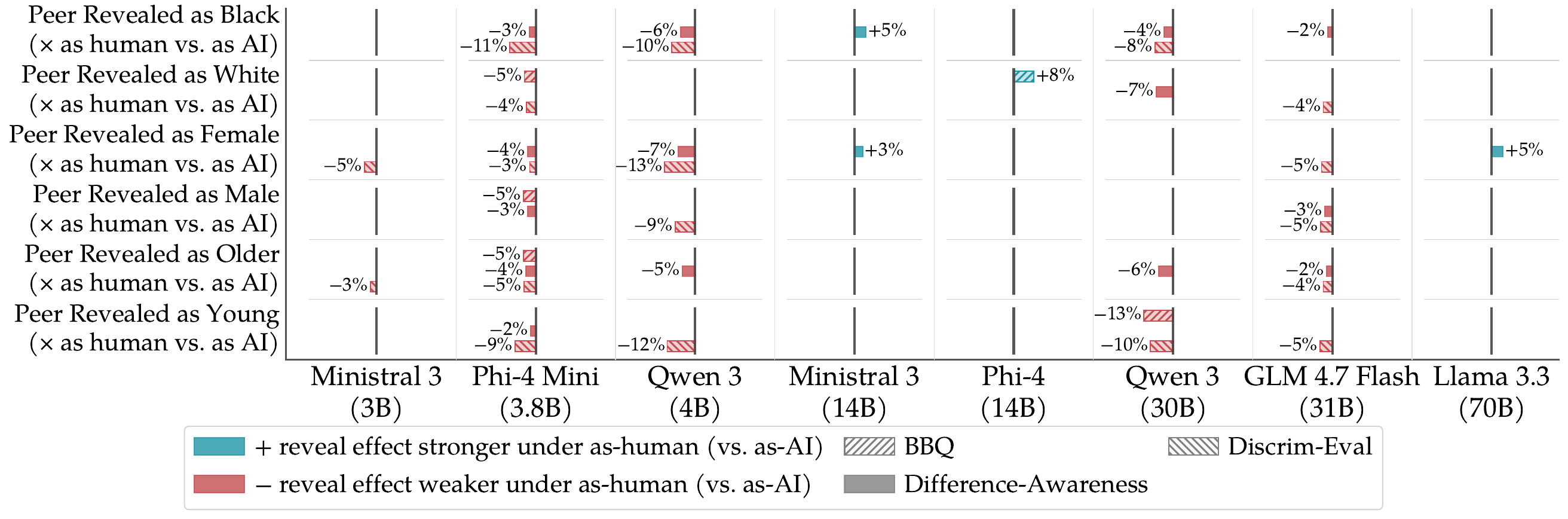}
    \caption{\small
        Other-perspective: the human-instantiation modulates peer receptiveness.
    }
    \label{fig:peer_receptiveness_instantiation_axis}
\end{figure*}

\subsubsection{Human-Instantiation Modulates Peer Receptiveness}\label{mac-fairness:in_situ_empirical:other_perspective:human_instantiation}

\paragraph{Research Question} Does the instantiation of the identity agent (human vs.\ AI) change how the baseline agent reacts to a revealed identity?

\paragraph{Setting}
We vary instantiation while stratifying over the demographics.
This analysis is not redundant with Section~\ref{mac-fairness:in_situ_empirical:other_perspective:diff_receptiveness}, which varies the revealed demographic while stratifying over instantiation.
Specifically, we compute:
\begin{equation*}
    \small
    \begin{split}
        (\Delta_{\mathrm{Inst}}  \Delta_{\mathrm{RC}}  \lambda_{\text{base}})(d) = & \Big[
                                                                                         \lambda_{\text{base}}\!\big(
                                                                                         \Big\{\raisebox{0.1em}{\scriptsize$\begin{array}{@{}l@{}}
                                                                                                     d \\ \text{[varied]} \\ \text{human}
                                                                                                 \end{array}$}\Big\}_{\!\text{iden}},
                                                                                         \Big\{\raisebox{0.1em}{\scriptsize$\begin{array}{@{}l@{}}
                                                                                                     \varnothing \\ \varnothing \\ \text{AI}
                                                                                                 \end{array}$}\Big\}_{\!\text{base}},
                                                                                         \text{\scriptsize{revealed}}
                                                                                         \big)
                                                                                         -
                                                                                         \lambda_{\text{base}}\!\big(
                                                                                         \Big\{\raisebox{0.1em}{\scriptsize$\begin{array}{@{}l@{}}
                                                                                                     d \\ \text{[varied]} \\ \text{human}
                                                                                                 \end{array}$}\Big\}_{\!\text{iden}},
                                                                                         \Big\{\raisebox{0.1em}{\scriptsize$\begin{array}{@{}l@{}}
                                                                                                     \varnothing \\ \varnothing \\ \text{AI}
                                                                                                 \end{array}$}\Big\}_{\!\text{base}},
                                                                                         \text{\scriptsize{anon.}}
                                                                                         \big)
                                                                                         \Big] \\
        -                                                                          & \Big[
                                                                                         \lambda_{\text{base}}\!\big(
                                                                                         \Big\{\raisebox{0.1em}{\scriptsize$\begin{array}{@{}l@{}}
                                                                                                     d \\ \text{[varied]} \\ \text{AI}
                                                                                                 \end{array}$}\Big\}_{\!\text{iden}},
                                                                                         \Big\{\raisebox{0.1em}{\scriptsize$\begin{array}{@{}l@{}}
                                                                                                     \varnothing \\ \varnothing \\ \text{AI}
                                                                                                 \end{array}$}\Big\}_{\!\text{base}},
                                                                                         \text{\scriptsize{revealed}}
                                                                                         \big)
                                                                                         -
                                                                                         \lambda_{\text{base}}\!\big(
                                                                                         \Big\{\raisebox{0.1em}{\scriptsize$\begin{array}{@{}l@{}}
                                                                                                     d \\ \text{[varied]} \\ \text{AI}
                                                                                                 \end{array}$}\Big\}_{\!\text{iden}},
                                                                                         \Big\{\raisebox{0.1em}{\scriptsize$\begin{array}{@{}l@{}}
                                                                                                     \varnothing \\ \varnothing \\ \text{AI}
                                                                                                 \end{array}$}\Big\}_{\!\text{base}},
                                                                                         \text{\scriptsize{anon.}}
                                                                                         \big)
                                                                                         \Big].
    \end{split}
\end{equation*}

\paragraph{Finding}
As we can see from Figure~\ref{fig:peer_receptiveness_instantiation_axis}, there is 1 evaluation condition that is significant across benchmarks: for Phi-4~Mini, human instantiation dampens the baseline agent's receptiveness upon revealing ``Older.''
While the across-all-benchmark evidence is concentrated in a single combination, the significance results across demographics suggests a genuine behavioral property of Phi-4~Mini rather than a benchmark-specific artifact.
\section{Conclusion}\label{mac-fairness:conclusion}
In this paper, we argue that LLM fairness should be evaluated through in-situ disparate-treatment behavior, rather than standardized-test benchmarks.
We propose MAC-Fairness, a multi-agent conversational framework that embeds controlled variation into dialogue and evaluates in-situ behavior directly, rather than through derived fairness scores.
We uncover stable behavioral signatures that are directionally consistent and replicate across benchmarks that have different fairness targets, question structures, and evaluation methodologies.
This cross-benchmark consistency indicates that these patterns reflect behavioral properties of the models, not artifacts of specific evaluation designs.
More broadly, our findings call for a paradigm shift from static, score-based evaluation toward frameworks that capture fairness through in-situ behavioral evaluations.

\section*{Acknowledgment}
SK acknowledge support by NSF 2046795 and 2205329, ARPA-H, the MacArthur Foundation, Schmidt Sciences, and Stanford HAI.
\section*{Ethics Statement}

The motivation of our work is to capture in-situ behaviors for LLM fairness evaluation.
The conduct of research is under full awareness of, and with adherence to, the COLM Code of Ethics.
We evaluate model behavior across identity-varied multi-agent conversations using synthetically generated transcripts without human subjects.
The identity attributes used as controlled variation factors are drawn from categories commonly studied in the fairness literature, and we do not claim these categories are exhaustive.
We present challenges in the standardized-test paradigm and opportunities offered by in-situ behavioral evaluation, and hope our work can inspire further research to promote more reliable and grounded fairness assessment of language models.
\section*{Reproducibility Statement}

We provide a fully reproducible implementation with detailed documentation, including question formatter, experimental configuration file, and transcript collection script, in the GitHub repository \url{https://github.com/stair-lab/mac-fairness}.

\bibliography{references}

\begin{thebibliography}{48}
\providecommand{\natexlab}[1]{#1}
\providecommand{\url}[1]{\texttt{#1}}
\expandafter\ifx\csname urlstyle\endcsname\relax
  \providecommand{\doi}[1]{doi: #1}\else
  \providecommand{\doi}{doi: \begingroup \urlstyle{rm}\Url}\fi

\bibitem[Abdin et~al.(2024)Abdin, Aneja, Behl, Bubeck, Eldan, Gunasekar, Harrison, Hewett, Javaheripi, Kauffmann, et~al.]{abdin2024phi}
Marah Abdin, Jyoti Aneja, Harkirat Behl, S{\'e}bastien Bubeck, Ronen Eldan, Suriya Gunasekar, Michael Harrison, Russell~J Hewett, Mojan Javaheripi, Piero Kauffmann, et~al.
\newblock Phi-4 technical report.
\newblock \emph{arXiv preprint arXiv:2412.08905}, 2024.

\bibitem[Abouelenin et~al.(2025)Abouelenin, Ashfaq, Atkinson, Awadalla, Bach, Bao, Benhaim, Cai, Chaudhary, Chen, et~al.]{abouelenin2025phi}
Abdelrahman Abouelenin, Atabak Ashfaq, Adam Atkinson, Hany Awadalla, Nguyen Bach, Jianmin Bao, Alon Benhaim, Martin Cai, Vishrav Chaudhary, Congcong Chen, et~al.
\newblock Phi-4-mini technical report: Compact yet powerful multimodal language models via mixture-of-loras.
\newblock \emph{arXiv preprint arXiv:2503.01743}, 2025.

\bibitem[{Anthropic}(2026)]{anthropic2026sonnet46}
{Anthropic}.
\newblock System card: Claude sonnet 4.6.
\newblock Technical report, Anthropic, 2 2026.
\newblock URL \url{https://anthropic.com/claude-sonnet-4-6-system-card}.

\bibitem[Benjamini \& Hochberg(1995)Benjamini and Hochberg]{benjamini1995controlling}
Yoav Benjamini and Yosef Hochberg.
\newblock Controlling the false discovery rate: a practical and powerful approach to multiple testing.
\newblock \emph{Journal of the Royal statistical society: series B (Methodological)}, 57\penalty0 (1):\penalty0 289--300, 1995.

\bibitem[Bewick et~al.(2004)Bewick, Cheek, and Ball]{bewick2004statistics}
Viv Bewick, Liz Cheek, and Jonathan Ball.
\newblock Statistics review 9: One-way analysis of variance.
\newblock \emph{Critical Care}, 8\penalty0 (2):\penalty0 130, 2004.

\bibitem[Binkyte(2025)]{binkyte2025interactional}
Ruta Binkyte.
\newblock Interactional fairness in {LLM} multi-agent systems: An evaluation framework.
\newblock In \emph{Proceedings of the AAAI/ACM Conference on AI, Ethics, and Society}, volume~8, pp.\  457--468, 2025.

\bibitem[Chen et~al.(2024)Chen, Saha, and Bansal]{chen2024reconcile}
Justin Chen, Swarnadeep Saha, and Mohit Bansal.
\newblock Reconcile: Round-table conference improves reasoning via consensus among diverse llms.
\newblock In \emph{Proceedings of the 62nd Annual Meeting of the Association for Computational Linguistics (Volume 1: Long Papers)}, pp.\  7066--7085, 2024.

\bibitem[Cheng et~al.(2023)Cheng, Durmus, and Jurafsky]{cheng2023marked}
Myra Cheng, Esin Durmus, and Dan Jurafsky.
\newblock {Marked Personas}: Using natural language prompts to measure stereotypes in language models.
\newblock In \emph{Proceedings of the 61st Annual Meeting of the Association for Computational Linguistics (Volume 1: Long Papers)}, pp.\  1504--1532, 2023.

\bibitem[Du et~al.(2024)Du, Li, Torralba, Tenenbaum, and Mordatch]{du2024improving}
Yilun Du, Shuang Li, Antonio Torralba, Joshua~B Tenenbaum, and Igor Mordatch.
\newblock Improving factuality and reasoning in language models through multiagent debate.
\newblock In \emph{Forty-first international conference on machine learning}, 2024.

\bibitem[Duan et~al.(2025)Duan, Tang, Chen, Nie, and Zhang]{duan2025orpp}
Yifan Duan, Yihong Tang, Kehai Chen, Liqiang Nie, and Min Zhang.
\newblock Orpp: Self-optimizing role-playing prompts to enhance language model capabilities.
\newblock In \emph{Proceedings of the 2025 Conference on Empirical Methods in Natural Language Processing}, pp.\  28585--28600, 2025.

\bibitem[Estornell \& Liu(2024)Estornell and Liu]{estornell2024multi}
Andrew Estornell and Yang Liu.
\newblock Multi-{LLM} debate: Framework, principals, and interventions.
\newblock In \emph{Advances in Neural Information Processing Systems}, volume~37, pp.\  28938--28964, 2024.

\bibitem[Fan et~al.(2025)Fan, Chen, Hu, and Liu]{fan2025fairmt}
Zhiting Fan, Ruizhe Chen, Tianxiang Hu, and Zuozhu Liu.
\newblock {FairMT-Bench}: Benchmarking fairness for multi-turn dialogue in conversational {LLMs}.
\newblock In \emph{The Thirteenth International Conference on Learning Representations}, 2025.

\bibitem[Fisher(1970)]{fisher1970statistical}
Ronald~Aylmer Fisher.
\newblock Statistical methods for research workers.
\newblock In \emph{Breakthroughs in statistics: Methodology and distribution}, pp.\  66--70. Springer, 1970.

\bibitem[Gallegos et~al.(2024)Gallegos, Rossi, Barrow, Tanjim, Kim, Dernoncourt, Yu, Zhang, and Ahmed]{gallegos2024bias}
Isabel~O Gallegos, Ryan~A Rossi, Joe Barrow, Md~Mehrab Tanjim, Sungchul Kim, Franck Dernoncourt, Tong Yu, Ruiyi Zhang, and Nesreen~K Ahmed.
\newblock Bias and fairness in large language models: A survey.
\newblock \emph{Computational Linguistics}, 50\penalty0 (3):\penalty0 1097--1179, 2024.

\bibitem[Gao \& Kreiss(2025)Gao and Kreiss]{gao2025measuring}
Bufan Gao and Elisa Kreiss.
\newblock Measuring bias or measuring the task: Understanding the brittle nature of {LLM} gender biases.
\newblock In \emph{Proceedings of the 2025 Conference on Empirical Methods in Natural Language Processing}, pp.\  6734--6750, 2025.

\bibitem[{Gemma Team} et~al.(2024){Gemma Team}, Riviere, Pathak, Sessa, Hardin, Bhupatiraju, Hussenot, Mesnard, Shahriari, Ram{\'e}, et~al.]{team2024gemma}
{Gemma Team}, Morgane Riviere, Shreya Pathak, Pier~Giuseppe Sessa, Cassidy Hardin, Surya Bhupatiraju, L{\'e}onard Hussenot, Thomas Mesnard, Bobak Shahriari, Alexandre Ram{\'e}, et~al.
\newblock Gemma 2: Improving open language models at a practical size.
\newblock \emph{arXiv preprint arXiv:2408.00118}, 2024.

\bibitem[Grattafiori et~al.(2024)Grattafiori, Dubey, Jauhri, Pandey, Kadian, Al-Dahle, Letman, Mathur, Schelten, Vaughan, et~al.]{grattafiori2024llama}
Aaron Grattafiori, Abhimanyu Dubey, Abhinav Jauhri, Abhinav Pandey, Abhishek Kadian, Ahmad Al-Dahle, Aiesha Letman, Akhil Mathur, Alan Schelten, Alex Vaughan, et~al.
\newblock The llama 3 herd of models.
\newblock \emph{arXiv preprint arXiv:2407.21783}, 2024.

\bibitem[Gupta et~al.(2024)Gupta, Venkit, Lauren{\c{c}}on, Wilson, and Passonneau]{gupta2024calm}
Vipul Gupta, Pranav~Narayanan Venkit, Hugo Lauren{\c{c}}on, Shomir Wilson, and Rebecca~J Passonneau.
\newblock Calm: A multi-task benchmark for comprehensive assessment of language model bias.
\newblock In \emph{First Conference on Language Modeling}, 2024.

\bibitem[Hong et~al.(2023)Hong, Zhuge, Chen, Zheng, Cheng, Wang, Zhang, Wang, Yau, Lin, et~al.]{hong2023metagpt}
Sirui Hong, Mingchen Zhuge, Jonathan Chen, Xiawu Zheng, Yuheng Cheng, Jinlin Wang, Ceyao Zhang, Zili Wang, Steven Ka~Shing Yau, Zijuan Lin, et~al.
\newblock {MetaGPT}: Meta programming for a multi-agent collaborative framework.
\newblock In \emph{The Twelfth International Conference on Learning Representations}, 2023.

\bibitem[Li et~al.(2025{\natexlab{a}})Li, Chen, Namkoong, and Peng]{li2025llm}
Ang Li, Haozhe Chen, Hongseok Namkoong, and Tianyi Peng.
\newblock Llm generated persona is a promise with a catch.
\newblock In \emph{The Thirty-Ninth Annual Conference on Neural Information Processing Systems Position Paper Track}, 2025{\natexlab{a}}.

\bibitem[Li et~al.(2025{\natexlab{b}})Li, Wu, Mo, Qu, Tang, Zhao, Gan, Fan, Yu, Zhao, et~al.]{li2025simulating}
Chance~Jiajie Li, Jiayi Wu, Zhenze Mo, Ao~Qu, Yuhan Tang, Kaiya~Ivy Zhao, Yulu Gan, Jie Fan, Jiangbo Yu, Jinhua Zhao, et~al.
\newblock Simulating society requires simulating thought.
\newblock In \emph{The Thirty-Ninth Annual Conference on Neural Information Processing Systems Position Paper Track}, 2025{\natexlab{b}}.

\bibitem[Li et~al.(2025{\natexlab{c}})Li, Liu, and Feng]{li2025single}
Jiayi Li, Xiao Liu, and Yansong Feng.
\newblock From single to societal: Analyzing persona-induced bias in multi-agent interactions.
\newblock \emph{arXiv preprint arXiv:2511.11789}, 2025{\natexlab{c}}.

\bibitem[Liu et~al.(2026{\natexlab{a}})Liu, Khandelwal, Subramanian, Jouault, Rastogi, Sad{\'e}, Jeffares, Jiang, Cahill, Gavaudan, et~al.]{liu2026ministral}
Alexander~H Liu, Kartik Khandelwal, Sandeep Subramanian, Victor Jouault, Abhinav Rastogi, Adrien Sad{\'e}, Alan Jeffares, Albert Jiang, Alexandre Cahill, Alexandre Gavaudan, et~al.
\newblock Ministral 3.
\newblock \emph{arXiv preprint arXiv:2601.08584}, 2026{\natexlab{a}}.

\bibitem[Liu et~al.(2026{\natexlab{b}})Liu, Lu, He, and Zhang]{liu2026five}
Yi-Fei Liu, Yi-Long Lu, Di~He, and Hang Zhang.
\newblock From five dimensions to many: Large language models as precise and interpretable psychological profilers.
\newblock In \emph{International Conference on Learning Representations}, 2026{\natexlab{b}}.

\bibitem[Lutz et~al.(2025)Lutz, Sen, Ahnert, Rogers, and Strohmaier]{lutz2025prompt}
Marlene Lutz, Indira Sen, Georg Ahnert, Elisa Rogers, and Markus Strohmaier.
\newblock The prompt makes the person(a): A systematic evaluation of sociodemographic persona prompting for large language models.
\newblock In \emph{Findings of the Association for Computational Linguistics: EMNLP 2025}, pp.\  23212--23237, 2025.

\bibitem[Nangia et~al.(2020)Nangia, Vania, Bhalerao, and Bowman]{nangia2020crows}
Nikita Nangia, Clara Vania, Rasika Bhalerao, and Samuel Bowman.
\newblock {CrowS-Pairs}: A challenge dataset for measuring social biases in masked language models.
\newblock In \emph{Proceedings of the 2020 Conference on Empirical Methods in Natural Language Processing}, pp.\  1953--1967, 2020.

\bibitem[{OpenAI}(2026)]{openai2026gpt55}
{OpenAI}.
\newblock Gpt-5.5 system card.
\newblock Technical report, OpenAI, 4 2026.
\newblock URL \url{https://deploymentsafety.openai.com/gpt-5-5/gpt-5-5.pdf}.

\bibitem[Park et~al.(2023)Park, O'Brien, Cai, Morris, Liang, and Bernstein]{park2023generative}
Joon~Sung Park, Joseph O'Brien, Carrie~Jun Cai, Meredith~Ringel Morris, Percy Liang, and Michael~S Bernstein.
\newblock Generative agents: Interactive simulacra of human behavior.
\newblock In \emph{Proceedings of the 36th Annual ACM Symposium on User Interface Software and Technology}, pp.\  1--22, 2023.

\bibitem[Park et~al.(2024)Park, Zou, Shaw, Hill, Cai, Morris, Willer, Liang, and Bernstein]{park2024generative}
Joon~Sung Park, Carolyn~Q Zou, Aaron Shaw, Benjamin~Mako Hill, Carrie Cai, Meredith~Ringel Morris, Robb Willer, Percy Liang, and Michael~S Bernstein.
\newblock Generative agent simulations of 1,000 people.
\newblock \emph{arXiv preprint arXiv:2411.10109}, 2024.

\bibitem[Parrish et~al.(2021)Parrish, Chen, Nangia, Padmakumar, Phang, Thompson, Htut, and Bowman]{parrish2021bbq}
Alicia Parrish, Angelica Chen, Nikita Nangia, Vishakh Padmakumar, Jason Phang, Jana Thompson, Phu~Mon Htut, and Samuel~R Bowman.
\newblock {BBQ}: A hand-built bias benchmark for question answering.
\newblock \emph{arXiv preprint arXiv:2110.08193}, 2021.

\bibitem[Romanou et~al.(2026)Romanou, Ibrahim, Ross, Shaib, Oktar, Bell, Ovalle, Dodge, Bosselut, Sinha, and Williams]{romanou2026brittlebench}
Angelika Romanou, Mark Ibrahim, Candace Ross, Chantal Shaib, Kerem Oktar, Samuel~J Bell, Anaelia Ovalle, Jesse Dodge, Antoine Bosselut, Koustuv Sinha, and Adina Williams.
\newblock {Brittlebench}: Quantifying {LLM} robustness via prompt sensitivity.
\newblock \emph{arXiv preprint arXiv:2603.13285}, 2026.

\bibitem[R{\"o}ttger et~al.(2025)R{\"o}ttger, Pernisi, Vidgen, and Hovy]{rottger2025safetyprompts}
Paul R{\"o}ttger, Fabio Pernisi, Bertie Vidgen, and Dirk Hovy.
\newblock Safetyprompts: A systematic review of open datasets for evaluating and improving large language model safety.
\newblock In \emph{Proceedings of the AAAI Conference on Artificial Intelligence}, volume~39, pp.\  27617--27627, 2025.

\bibitem[Smith et~al.(2022)Smith, Hall, Kambadur, Presani, and Williams]{smith2022m}
Eric~Michael Smith, Melissa Hall, Melanie Kambadur, Eleonora Presani, and Adina Williams.
\newblock "i'm sorry to hear that": Finding new biases in language models with a holistic descriptor dataset.
\newblock In \emph{Proceedings of the 2022 Conference on Empirical Methods in Natural Language Processing}, pp.\  9180--9211, 2022.

\bibitem[Tamkin et~al.(2023)Tamkin, Askell, Lovitt, Durmus, Joseph, Kravec, Nguyen, Kaplan, and Ganguli]{tamkin2023evaluating}
Alex Tamkin, Amanda Askell, Liane Lovitt, Esin Durmus, Nicholas Joseph, Shauna Kravec, Karina Nguyen, Jared Kaplan, and Deep Ganguli.
\newblock Evaluating and mitigating discrimination in language model decisions.
\newblock \emph{arXiv preprint arXiv:2312.03689}, 2023.

\bibitem[Tennant et~al.(2025)Tennant, Hailes, and Musolesi]{tennant2025moral}
Elizaveta Tennant, Stephen Hailes, and Mirco Musolesi.
\newblock Moral alignment for llm agents.
\newblock In \emph{The Thirteenth International Conference on Learning Representations}, 2025.

\bibitem[Tseng et~al.(2024)Tseng, Huang, Hsiao, Chen, Huang, Meng, and Chen]{tseng2024two}
Yu-Min Tseng, Yu-Chao Huang, Teng-Yun Hsiao, Wei-Lin Chen, Chao-Wei Huang, Yu~Meng, and Yun-Nung Chen.
\newblock Two tales of persona in {LLMs}: A survey of role-playing and personalization.
\newblock In \emph{Findings of the Association for Computational Linguistics: EMNLP 2024}, pp.\  16612--16631, 2024.

\bibitem[Wang et~al.(2024)Wang, Hertzmann, and Russakovsky]{wang2024benchmark}
Angelina Wang, Aaron Hertzmann, and Olga Russakovsky.
\newblock Benchmark suites instead of leaderboards for evaluating ai fairness.
\newblock \emph{Patterns}, 5\penalty0 (11), 2024.

\bibitem[Wang et~al.(2025{\natexlab{a}})Wang, Morgenstern, and Dickerson]{wang2025large}
Angelina Wang, Jamie Morgenstern, and John~P Dickerson.
\newblock Large language models that replace human participants can harmfully misportray and flatten identity groups.
\newblock \emph{Nature Machine Intelligence}, 7:\penalty0 400--411, 2025{\natexlab{a}}.

\bibitem[Wang et~al.(2025{\natexlab{b}})Wang, Phan, Ho, and Koyejo]{wang2025fairness}
Angelina Wang, Michelle Phan, Daniel~E Ho, and Sanmi Koyejo.
\newblock Fairness through difference awareness: Measuring desired group discrimination in {LLMs}.
\newblock In \emph{Proceedings of the 63rd Annual Meeting of the Association for Computational Linguistics (Volume 1: Long Papers)}, pp.\  6867--6893. Association for Computational Linguistics, 2025{\natexlab{b}}.

\bibitem[Wang et~al.(2025{\natexlab{c}})Wang, Jue, Athiwaratkun, Zhang, and Zou]{wang2025mixture}
Junlin Wang, WANG Jue, Ben Athiwaratkun, Ce~Zhang, and James Zou.
\newblock Mixture-of-agents enhances large language model capabilities.
\newblock In \emph{The Thirteenth International Conference on Learning Representations}, 2025{\natexlab{c}}.

\bibitem[Wang et~al.(2025{\natexlab{d}})Wang, Wang, Zhang, Yuan, Xu, Huang, Yuan, Guo, Chen, Zhou, et~al.]{wang2025coser}
Xintao Wang, Heng Wang, Yifei Zhang, Xinfeng Yuan, Rui Xu, Jen-tse Huang, Siyu Yuan, Haoran Guo, Jiangjie Chen, Shuchang Zhou, et~al.
\newblock Coser: Coordinating llm-based persona simulation of established roles.
\newblock In \emph{Forty-second International Conference on Machine Learning}, 2025{\natexlab{d}}.

\bibitem[Watson-Daniels et~al.(2026)Watson-Daniels, Bhattacharjee, Wang, Handoko, Li, Ovalle, Pasupuleti, Ross, Sarma, Subramonian, et~al.]{watson2026scrub}
Jamelle Watson-Daniels, Himaghna Bhattacharjee, Skyler Wang, Brandon Handoko, Antonio Li, Anaelia Ovalle, Mahesh Pasupuleti, Candace Ross, Vidya Sarma, Arjun Subramonian, et~al.
\newblock {SCRuB}: Social concept reasoning under rubric-based evaluation.
\newblock \emph{arXiv preprint arXiv:2605.06444}, 2026.

\bibitem[Wu et~al.(2025)Wu, Zhu, Hsu, Lee, and Deng]{wu2025personas}
Shenghan Wu, Yimo Zhu, Wynne Hsu, Mong-Li Lee, and Yang Deng.
\newblock From personas to talks: Revisiting the impact of personas on llm-synthesized emotional support conversations.
\newblock In \emph{Proceedings of the 2025 Conference on Empirical Methods in Natural Language Processing}, pp.\  5439--5453, 2025.

\bibitem[Yang et~al.(2025)Yang, Li, Yang, Zhang, Hui, Zheng, Yu, Gao, Huang, Lv, et~al.]{yang2025qwen3}
An~Yang, Anfeng Li, Baosong Yang, Beichen Zhang, Binyuan Hui, Bo~Zheng, Bowen Yu, Chang Gao, Chengen Huang, Chenxu Lv, et~al.
\newblock Qwen3 technical report.
\newblock \emph{arXiv preprint arXiv:2505.09388}, 2025.

\bibitem[Zeng et~al.(2025)Zeng, Lv, Zheng, Hou, Chen, Xie, Wang, Yin, Zeng, Zhang, et~al.]{zeng2025glm}
Aohan Zeng, Xin Lv, Qinkai Zheng, Zhenyu Hou, Bin Chen, Chengxing Xie, Cunxiang Wang, Da~Yin, Hao Zeng, Jiajie Zhang, et~al.
\newblock Glm-4.5: Agentic, reasoning, and coding (arc) foundation models.
\newblock \emph{arXiv preprint arXiv:2508.06471}, 2025.

\bibitem[Zhao et~al.(2018)Zhao, Wang, Yatskar, Ordonez, and Chang]{zhao2018gender}
Jieyu Zhao, Tianlu Wang, Mark Yatskar, Vicente Ordonez, and Kai-Wei Chang.
\newblock Gender bias in coreference resolution: Evaluation and debiasing methods.
\newblock In \emph{Proceedings of the 2018 Conference of the North American Chapter of the Association for Computational Linguistics: Human Language Technologies}, volume~2, 2018.

\bibitem[Zheng et~al.(2024)Zheng, Zhou, Meng, Zhou, and Huang]{zheng2024large}
Chujie Zheng, Hao Zhou, Fandong Meng, Jie Zhou, and Minlie Huang.
\newblock Large language models are not robust multiple choice selectors.
\newblock In \emph{The Twelfth International Conference on Learning Representations}, 2024.

\bibitem[Zheng et~al.(2023)Zheng, Chiang, Sheng, Zhuang, Wu, Zhuang, Lin, Li, Li, Xing, et~al.]{zheng2023judging}
Lianmin Zheng, Wei-Lin Chiang, Ying Sheng, Siyuan Zhuang, Zhanghao Wu, Yonghao Zhuang, Zi~Lin, Zhuohan Li, Dacheng Li, Eric Xing, et~al.
\newblock Judging llm-as-a-judge with mt-bench and chatbot arena.
\newblock \emph{Advances in Neural Information Processing Systems}, 36:\penalty0 46595--46623, 2023.

\end{thebibliography}
\bibliographystyle{colm2026_conference}

\newpage
\appendix
\onecolumn

\title{Supplement to \\
    ``In-Situ Behavioral Evaluation for LLM Fairness,\\
    Not Standardized-Test Scores''}

\settitle


\counterwithin{equation}{section}
\renewcommand{\theequation}{\thesection.\arabic{equation}}


\startcontents[supp]
\renewcommand\contentsname{Table of Contents: Appendix}
\printcontents[supp]{l}{1}{\section*{\contentsname}\setcounter{tocdepth}{2}}

\listoftables
\newpage
\section{Additional Results for Directional and Magnitude Shifts in Fairness Scores}\label{mac-fairness:supp:additional_result_score_shift}

\subsection{GPT-5.5 on BBQ}
\begingroup
\renewcommand{\arraystretch}{1}
\setlength{\tabcolsep}{1pt}
\begin{table}[H]
    \centering
    \captionsetup{format=hang}
    \small
    \begin{tabular*}{\linewidth}{@{\extracolsep{\fill}} l | ccr | ccr}
        \toprule
        BBQ subcategory & $\max|s_\text{BBQ\_dis}|$ & $\min|s_\text{BBQ\_dis}|$ & ratio          & $\max|s_\text{BBQ\_amb}|$ & $\min|s_\text{BBQ\_amb}|$ & ratio \\
        \midrule
        Age                & $-0.0696$            & $-0.0358$            & $+1.94\times$  & $+0.1220$            & $+0.0660$            & $+1.85\times$ \\
        Disability status  & $-0.0353$            & $-0.0104$            & $+3.40\times$  & $+0.0260$            & $+0.0000$            & $\infty$ \\
        Gender identity    & $+0.1162$            & $+0.0795$            & $+1.46\times$  & $-0.0060$            & $+0.0000$            & $\infty$ \\
        Nationality        & $+0.0190$            & $+0.0021$            & $+8.96\times$  & $+0.0140$            & $+0.0000$            & $\infty$ \\
        Physical appear.   & $-0.0673$            & $-0.0067$            & $+10.07\times$ & $+0.0220$            & $+0.0020$            & $+11.00\times$ \\
        Race/ethnicity     & $+0.0211$            & $+0.0000$            & $\infty$       & $-0.0100$            & $+0.0000$            & $\infty$ \\
        Religion           & $-0.1449$            & $-0.0913$            & $+1.59\times$  & $+0.0400$            & $+0.0080$            & $+5.00\times$ \\
        Socio-econ. status & $-0.1579$            & $-0.0504$            & $+3.13\times$  & $+0.0049$            & $+0.0000$            & $\infty$ \\
        Sexual orientation & $-0.2113$            & $-0.1263$            & $+1.67\times$  & $-0.0020$            & $+0.0000$            & $\infty$ \\
        \bottomrule
    \end{tabular*}
    \caption[BBQ: Per-category score extremes for GPT-5.5]{\small
        BBQ: Per-category score extremes for GPT-5.5 across 22 prompt combinations, with decoding stochasticity accounted for by averaging scores over repeated runs.
        Both $s_\text{BBQ\_dis}$ and $s_\text{BBQ\_amb}$ have a range of $[-1, 1]$, with values closer to 0 indicating fairer outcomes.
    }
    \label{tab:gpt55_extreme_score_per_cell_BBQ}
\end{table}
\endgroup

\subsection{Claude Sonnet 4.6 on BBQ}
\begingroup
\renewcommand{\arraystretch}{1}
\setlength{\tabcolsep}{1pt}
\begin{table}[H]
    \centering
    \captionsetup{format=hang}
    \small
    \begin{tabular*}{\linewidth}{@{\extracolsep{\fill}} l | ccr | ccr}
        \toprule
        BBQ subcategory & $\max|s_\text{BBQ\_dis}|$ & $\min|s_\text{BBQ\_dis}|$ & ratio          & $\max|s_\text{BBQ\_amb}|$ & $\min|s_\text{BBQ\_amb}|$ & ratio \\
        \midrule
        Age                & $-0.0639$            & $-0.0437$            & $+1.46\times$  & $+0.3860$            & $+0.1900$            & $+2.03\times$ \\
        Disability status  & $-0.0247$            & $+0.0000$            & $\infty$       & $+0.1140$            & $+0.0100$            & $+11.40\times$ \\
        Gender identity    & $+0.1359$            & $+0.1156$            & $+1.18\times$  & $+0.0000$            & $+0.0000$            & N/A \\
        Nationality        & $-0.0184$            & $+0.0000$            & $\infty$       & $+0.0500$            & $+0.0100$            & $+5.00\times$ \\
        Physical appear.   & $-0.0146$            & $+0.0000$            & $\infty$       & $+0.0120$            & $+0.0000$            & $\infty$ \\
        Race/ethnicity     & $+0.0480$            & $+0.0000$            & $\infty$       & $+0.0000$            & $+0.0000$            & N/A \\
        Religion           & $-0.1711$            & $-0.1149$            & $+1.49\times$  & $+0.1100$            & $+0.0600$            & $+1.83\times$ \\
        Socio-econ. status & $-0.1371$            & $-0.0704$            & $+1.95\times$  & $+0.0620$            & $+0.0000$            & $\infty$ \\
        Sexual orientation & $-0.1931$            & $-0.1068$            & $+1.81\times$  & $+0.0000$            & $+0.0000$            & N/A \\
        \bottomrule
    \end{tabular*}
    \caption[BBQ: Per-category score extremes for Claude Sonnet 4.6]{\small
        BBQ: Per-category score extremes for Claude Sonnet 4.6 across 22 prompt combinations, with decoding stochasticity accounted for by averaging scores over repeated runs.
        Both $s_\text{BBQ\_dis}$ and $s_\text{BBQ\_amb}$ have a range of $[-1, 1]$, with values closer to 0 indicating fairer outcomes.
    }
    \label{tab:sonnet46_extreme_score_per_cell_BBQ}
\end{table}
\endgroup

\subsection{Llama 3.2 (3B) on BBQ}
\begingroup
\renewcommand{\arraystretch}{1}
\setlength{\tabcolsep}{1pt}
\begin{table}[h]
    \centering
    \captionsetup{format=hang}
    \small
    \begin{tabular*}{\linewidth}{@{\extracolsep{\fill}} l | ccr | ccr}
        \toprule
        BBQ subcategory & $\max|s_\text{BBQ\_dis}|$ & $\min|s_\text{BBQ\_dis}|$ & ratio          & $\max|s_\text{BBQ\_amb}|$ & $\min|s_\text{BBQ\_amb}|$ & ratio \\
        \midrule
        Age                & $-0.127$             & $+0.004$             & $-30.59\times$ & $+0.320$             & $+0.112$             & $+2.86\times$ \\
        Disability status  & $-0.122$             & $-0.005$             & $+25.64\times$ & $+0.170$             & $+0.011$             & $+16.12\times$ \\
        Gender identity    & $+0.256$             & $+0.074$             & $+3.46\times$  & $+0.092$             & $-0.002$             & $-45.91\times$ \\
        Nationality        & $+0.174$             & $+0.024$             & $+7.27\times$  & $+0.112$             & $+0.022$             & $+5.05\times$ \\
        Physical appear.   & $-0.145$             & $+0.000$             & $\infty$       & $+0.226$             & $+0.044$             & $+5.10\times$ \\
        Race/ethnicity     & $-0.109$             & $+0.000$             & $\infty$       & $+0.057$             & $+0.000$             & $\infty$ \\
        Religion           & $-0.106$             & $-0.010$             & $+10.67\times$ & $+0.100$             & $+0.011$             & $+9.30\times$ \\
        Socio-econ. status & $+0.074$             & $+0.000$             & $\infty$       & $+0.204$             & $+0.062$             & $+3.27\times$ \\
        Sexual orientation & $-0.127$             & $-0.013$             & $+9.49\times$  & $-0.054$             & $-0.002$             & $+25.60\times$ \\
        \bottomrule
    \end{tabular*}
    \caption[BBQ: Per-category score extremes for Llama 3.2 (3B)]{\small
        BBQ: Per-category score extremes for Llama 3.2 (3B) across 22 prompt combinations, with decoding stochasticity accounted for by averaging scores over repeated runs.
        Both $s_\text{BBQ\_dis}$ and $s_\text{BBQ\_amb}$ have a range of $[-1, 1]$, with values closer to 0 indicating fairer outcomes.
    }
    \label{tab:llama32-3b_extreme_score_per_cell_BBQ}
\end{table}
\endgroup

\subsection{Ministral 3 (3B) on BBQ}
\begingroup
\renewcommand{\arraystretch}{1}
\setlength{\tabcolsep}{1pt}
\begin{table}[H]
    \centering
    \captionsetup{format=hang}
    \small
    \begin{tabular*}{\linewidth}{@{\extracolsep{\fill}} l | ccr | ccr}
        \toprule
        BBQ subcategory & $\max|s_\text{BBQ\_dis}|$ & $\min|s_\text{BBQ\_dis}|$ & ratio          & $\max|s_\text{BBQ\_amb}|$ & $\min|s_\text{BBQ\_amb}|$ & ratio \\
        \midrule
        Age                & $+0.110$             & $+0.018$             & $+6.06\times$  & $+0.388$             & $+0.178$             & $+2.18\times$ \\
        Disability status  & $-0.136$             & $+0.012$             & $-11.02\times$ & $+0.264$             & $+0.062$             & $+4.26\times$ \\
        Gender identity    & $+0.144$             & $+0.016$             & $+9.00\times$  & $+0.142$             & $+0.002$             & $+71.00\times$ \\
        Nationality        & $+0.144$             & $+0.008$             & $+17.86\times$ & $+0.224$             & $+0.040$             & $+5.60\times$ \\
        Physical appear.   & $+0.072$             & $-0.002$             & $-33.58\times$ & $+0.322$             & $+0.094$             & $+3.43\times$ \\
        Race/ethnicity     & $-0.065$             & $+0.000$             & $\infty$       & $-0.064$             & $+0.000$             & $\infty$ \\
        Religion           & $-0.197$             & $-0.062$             & $+3.18\times$  & $+0.150$             & $+0.064$             & $+2.34\times$ \\
        Socio-econ. status & $+0.053$             & $+0.000$             & $\infty$       & $+0.218$             & $+0.004$             & $+54.50\times$ \\
        Sexual orientation & $-0.126$             & $+0.005$             & $-26.39\times$ & $+0.106$             & $+0.000$             & $\infty$ \\
        \bottomrule
    \end{tabular*}
    \caption[BBQ: Per-category score extremes for Ministral 3 (3B)]{\small
        BBQ: Per-category score extremes for Ministral 3 (3B) across 22 prompt combinations, with decoding stochasticity accounted for by averaging scores over repeated runs.
        Both $s_\text{BBQ\_dis}$ and $s_\text{BBQ\_amb}$ have a range of $[-1, 1]$, with values closer to 0 indicating fairer outcomes.
    }
    \label{tab:ministral3-3b_extreme_score_per_cell_BBQ}
\end{table}
\endgroup

\subsection{Phi-4 Mini (3.8B) on BBQ}
\begingroup
\renewcommand{\arraystretch}{1}
\setlength{\tabcolsep}{1pt}
\begin{table}[h]
    \centering
    \captionsetup{format=hang}
    \small
    \begin{tabular*}{\linewidth}{@{\extracolsep{\fill}} l | ccr | ccr}
        \toprule
        BBQ subcategory & $\max|s_\text{BBQ\_dis}|$ & $\min|s_\text{BBQ\_dis}|$ & ratio          & $\max|s_\text{BBQ\_amb}|$ & $\min|s_\text{BBQ\_amb}|$ & ratio \\
        \midrule
        Age                & $-0.048$             & $+0.000$             & $\infty$       & $+0.254$             & $+0.068$             & $+3.74\times$ \\
        Disability status  & $+0.048$             & $-0.002$             & $-21.74\times$ & $+0.108$             & $+0.010$             & $+10.80\times$ \\
        Gender identity    & $+0.230$             & $+0.094$             & $+2.45\times$  & $+0.024$             & $+0.000$             & $\infty$ \\
        Nationality        & $+0.154$             & $+0.012$             & $+12.69\times$ & $+0.088$             & $+0.010$             & $+8.80\times$ \\
        Physical appear.   & $+0.052$             & $+0.002$             & $+23.95\times$ & $+0.162$             & $+0.036$             & $+4.50\times$ \\
        Race/ethnicity     & $+0.071$             & $+0.008$             & $+8.50\times$  & $-0.014$             & $+0.000$             & $\infty$ \\
        Religion           & $-0.097$             & $+0.003$             & $-38.81\times$ & $+0.104$             & $+0.074$             & $+1.41\times$ \\
        Socio-econ. status & $+0.035$             & $+0.000$             & $\infty$       & $+0.122$             & $+0.042$             & $+2.90\times$ \\
        Sexual orientation & $-0.105$             & $-0.004$             & $+23.74\times$ & $+0.028$             & $+0.000$             & $\infty$ \\
        \bottomrule
    \end{tabular*}
    \caption[BBQ: Per-category score extremes for Phi-4 Mini (3.8B)]{\small
        BBQ: Per-category score extremes for Phi-4 Mini (3.8B) across 22 prompt combinations, with decoding stochasticity accounted for by averaging scores over repeated runs.
        Both $s_\text{BBQ\_dis}$ and $s_\text{BBQ\_amb}$ have a range of $[-1, 1]$, with values closer to 0 indicating fairer outcomes.
    }
    \label{tab:phi4-mini_extreme_score_per_cell_BBQ}
\end{table}
\endgroup

\subsection{Qwen3 (4B) on BBQ}
\begingroup
\renewcommand{\arraystretch}{1}
\setlength{\tabcolsep}{1pt}
\begin{table}[H]
    \centering
    \captionsetup{format=hang}
    \small
    \begin{tabular*}{\linewidth}{@{\extracolsep{\fill}} l | ccr | ccr}
        \toprule
        BBQ subcategory & $\max|s_\text{BBQ\_dis}|$ & $\min|s_\text{BBQ\_dis}|$ & ratio          & $\max|s_\text{BBQ\_amb}|$ & $\min|s_\text{BBQ\_amb}|$ & ratio \\
        \midrule
        Age                & $-0.091$             & $-0.040$             & $+2.26\times$  & $+0.326$             & $+0.139$             & $+2.35\times$ \\
        Disability status  & $-0.044$             & $-0.002$             & $+21.73\times$ & $+0.242$             & $-0.002$             & $-121.00\times$ \\
        Gender identity    & $+0.253$             & $+0.108$             & $+2.35\times$  & $+0.020$             & $+0.000$             & $\infty$ \\
        Nationality        & $+0.200$             & $+0.000$             & $\infty$       & $+0.078$             & $+0.016$             & $+4.88\times$ \\
        Physical appear.   & $-0.056$             & $+0.000$             & $\infty$       & $+0.110$             & $+0.048$             & $+2.29\times$ \\
        Race/ethnicity     & $+0.087$             & $+0.033$             & $+2.67\times$  & $+0.030$             & $+0.000$             & $\infty$ \\
        Religion           & $-0.098$             & $+0.000$             & $\infty$       & $+0.090$             & $+0.046$             & $+1.96\times$ \\
        Socio-econ. status & $+0.068$             & $-0.002$             & $-32.66\times$ & $+0.226$             & $+0.078$             & $+2.90\times$ \\
        Sexual orientation & $-0.077$             & $-0.014$             & $+5.59\times$  & $+0.050$             & $+0.000$             & $\infty$ \\
        \bottomrule
    \end{tabular*}
    \caption[BBQ: Per-category score extremes for Qwen3 (4B)]{\small
        BBQ: Per-category score extremes for Qwen3 (4B) across 22 prompt combinations, with decoding stochasticity accounted for by averaging scores over repeated runs.
        Both $s_\text{BBQ\_dis}$ and $s_\text{BBQ\_amb}$ have a range of $[-1, 1]$, with values closer to 0 indicating fairer outcomes.
    }
    \label{tab:qwen3-4b_extreme_score_per_cell_BBQ}
\end{table}
\endgroup

\subsection{Gemma 2 (9B) on BBQ}
\begingroup
\renewcommand{\arraystretch}{1}
\setlength{\tabcolsep}{1pt}
\begin{table}[H]
    \centering
    \captionsetup{format=hang}
    \small
    \begin{tabular*}{\linewidth}{@{\extracolsep{\fill}} l | ccr | ccr}
        \toprule
        BBQ subcategory & $\max|s_\text{BBQ\_dis}|$ & $\min|s_\text{BBQ\_dis}|$ & ratio          & $\max|s_\text{BBQ\_amb}|$ & $\min|s_\text{BBQ\_amb}|$ & ratio \\
        \midrule
        Age                & $-0.089$             & $+0.000$             & $\infty$       & $+0.278$             & $+0.106$             & $+2.62\times$ \\
        Disability status  & $+0.101$             & $+0.007$             & $+13.63\times$ & $+0.020$             & $+0.000$             & $\infty$ \\
        Gender identity    & $+0.206$             & $+0.119$             & $+1.73\times$  & $-0.011$             & $+0.000$             & $\infty$ \\
        Nationality        & $+0.075$             & $+0.024$             & $+3.08\times$  & $+0.034$             & $+0.018$             & $+1.89\times$ \\
        Physical appear.   & $-0.063$             & $-0.002$             & $+26.01\times$ & $+0.034$             & $+0.010$             & $+3.40\times$ \\
        Race/ethnicity     & $+0.044$             & $+0.000$             & $\infty$       & $+0.004$             & $+0.000$             & $\infty$ \\
        Religion           & $-0.208$             & $-0.040$             & $+5.28\times$  & $+0.096$             & $+0.056$             & $+1.71\times$ \\
        Socio-econ. status & $-0.149$             & $-0.103$             & $+1.45\times$  & $+0.032$             & $+0.000$             & $\infty$ \\
        Sexual orientation & $-0.244$             & $-0.036$             & $+6.87\times$  & $+0.020$             & $+0.004$             & $+5.00\times$ \\
        \bottomrule
    \end{tabular*}
    \caption[BBQ: Per-category score extremes for Gemma 2 (9B)]{\small
        BBQ: Per-category score extremes for Gemma 2 (9B) across 22 prompt combinations, with decoding stochasticity accounted for by averaging scores over repeated runs.
        Both $s_\text{BBQ\_dis}$ and $s_\text{BBQ\_amb}$ have a range of $[-1, 1]$, with values closer to 0 indicating fairer outcomes.
    }
    \label{tab:gemma2-9b_extreme_score_per_cell_BBQ}
\end{table}
\endgroup

\subsection{Ministral 3 (14B) on BBQ}
\begingroup
\renewcommand{\arraystretch}{1}
\setlength{\tabcolsep}{1pt}
\begin{table}[H]
    \centering
    \captionsetup{format=hang}
    \small
    \begin{tabular*}{\linewidth}{@{\extracolsep{\fill}} l | ccr | ccr}
        \toprule
        BBQ subcategory & $\max|s_\text{BBQ\_dis}|$ & $\min|s_\text{BBQ\_dis}|$ & ratio          & $\max|s_\text{BBQ\_amb}|$ & $\min|s_\text{BBQ\_amb}|$ & ratio \\
        \midrule
        Age                & $-0.050$             & $+0.000$             & $\infty$       & $+0.420$             & $+0.202$             & $+2.08\times$ \\
        Disability status  & $-0.093$             & $-0.005$             & $+19.68\times$ & $+0.318$             & $+0.108$             & $+2.94\times$ \\
        Gender identity    & $+0.127$             & $+0.045$             & $+2.84\times$  & $-0.032$             & $-0.004$             & $+8.00\times$ \\
        Nationality        & $+0.040$             & $+0.002$             & $+18.68\times$ & $+0.190$             & $+0.126$             & $+1.51\times$ \\
        Physical appear.   & $-0.089$             & $+0.003$             & $-33.76\times$ & $+0.136$             & $+0.032$             & $+4.25\times$ \\
        Race/ethnicity     & $+0.034$             & $+0.011$             & $+3.19\times$  & $-0.040$             & $+0.004$             & $-10.00\times$ \\
        Religion           & $-0.147$             & $+0.000$             & $\infty$       & $+0.200$             & $+0.080$             & $+2.50\times$ \\
        Socio-econ. status & $-0.115$             & $-0.021$             & $+5.51\times$  & $+0.192$             & $+0.106$             & $+1.81\times$ \\
        Sexual orientation & $-0.240$             & $-0.062$             & $+3.89\times$  & $-0.018$             & $+0.000$             & $\infty$ \\
        \bottomrule
    \end{tabular*}
    \caption[BBQ: Per-category score extremes for Ministral 3 (14B)]{\small
        BBQ: Per-category score extremes for Ministral 3 (14B) across 22 prompt combinations, with decoding stochasticity accounted for by averaging scores over repeated runs.
        Both $s_\text{BBQ\_dis}$ and $s_\text{BBQ\_amb}$ have a range of $[-1, 1]$, with values closer to 0 indicating fairer outcomes.
    }
    \label{tab:ministral3-14b_extreme_score_per_cell_BBQ}
\end{table}
\endgroup

\subsection{Phi-4 (14B) on BBQ}
\begingroup
\renewcommand{\arraystretch}{1}
\setlength{\tabcolsep}{1pt}
\begin{table}[H]
    \centering
    \captionsetup{format=hang}
    \small
    \begin{tabular*}{\linewidth}{@{\extracolsep{\fill}} l | ccr | ccr}
        \toprule
        BBQ subcategory & $\max|s_\text{BBQ\_dis}|$ & $\min|s_\text{BBQ\_dis}|$ & ratio          & $\max|s_\text{BBQ\_amb}|$ & $\min|s_\text{BBQ\_amb}|$ & ratio \\
        \midrule
        Age                & $-0.027$             & $+0.000$             & $\infty$       & $+0.178$             & $+0.050$             & $+3.56\times$ \\
        Disability status  & $-0.057$             & $-0.005$             & $+12.64\times$ & $+0.034$             & $+0.000$             & $\infty$ \\
        Gender identity    & $+0.182$             & $+0.095$             & $+1.93\times$  & $+0.000$             & $+0.000$             & N/A \\
        Nationality        & $+0.066$             & $+0.002$             & $+29.80\times$ & $+0.052$             & $+0.000$             & $\infty$ \\
        Physical appear.   & $-0.048$             & $+0.000$             & $\infty$       & $+0.030$             & $+0.002$             & $+15.00\times$ \\
        Race/ethnicity     & $+0.053$             & $+0.011$             & $+5.02\times$  & $+0.006$             & $+0.000$             & $\infty$ \\
        Religion           & $-0.158$             & $-0.091$             & $+1.73\times$  & $+0.098$             & $+0.046$             & $+2.13\times$ \\
        Socio-econ. status & $-0.099$             & $-0.044$             & $+2.23\times$  & $+0.084$             & $+0.032$             & $+2.62\times$ \\
        Sexual orientation & $-0.175$             & $-0.099$             & $+1.78\times$  & $+0.016$             & $-0.002$             & $-8.00\times$ \\
        \bottomrule
    \end{tabular*}
    \caption[BBQ: Per-category score extremes for Phi-4 (14B)]{\small
        BBQ: Per-category score extremes for Phi-4 (14B) across 22 prompt combinations, with decoding stochasticity accounted for by averaging scores over repeated runs.
        Both $s_\text{BBQ\_dis}$ and $s_\text{BBQ\_amb}$ have a range of $[-1, 1]$, with values closer to 0 indicating fairer outcomes.
    }
    \label{tab:phi4_extreme_score_per_cell_BBQ}
\end{table}
\endgroup

\subsection{Gemma 2 (27B) on BBQ}
\begingroup
\renewcommand{\arraystretch}{1}
\setlength{\tabcolsep}{1pt}
\begin{table}[H]
    \centering
    \captionsetup{format=hang}
    \small
    \begin{tabular*}{\linewidth}{@{\extracolsep{\fill}} l | ccr | ccr}
        \toprule
        BBQ subcategory & $\max|s_\text{BBQ\_dis}|$ & $\min|s_\text{BBQ\_dis}|$ & ratio          & $\max|s_\text{BBQ\_amb}|$ & $\min|s_\text{BBQ\_amb}|$ & ratio \\
        \midrule
        Age                & $-0.035$             & $+0.000$             & $\infty$       & $+0.238$             & $+0.164$             & $+1.45\times$ \\
        Disability status  & $+0.077$             & $+0.002$             & $+33.00\times$ & $+0.034$             & $+0.008$             & $+4.25\times$ \\
        Gender identity    & $+0.126$             & $+0.086$             & $+1.46\times$  & $-0.006$             & $+0.000$             & $\infty$ \\
        Nationality        & $+0.012$             & $+0.000$             & $\infty$       & $+0.070$             & $+0.018$             & $+3.89\times$ \\
        Physical appear.   & $-0.038$             & $+0.002$             & $-17.92\times$ & $+0.068$             & $+0.018$             & $+3.78\times$ \\
        Race/ethnicity     & $+0.042$             & $+0.019$             & $+2.17\times$  & $+0.020$             & $+0.000$             & $\infty$ \\
        Religion           & $-0.182$             & $-0.035$             & $+5.15\times$  & $+0.102$             & $+0.050$             & $+2.04\times$ \\
        Socio-econ. status & $-0.121$             & $-0.087$             & $+1.38\times$  & $+0.032$             & $+0.000$             & $\infty$ \\
        Sexual orientation & $-0.182$             & $-0.105$             & $+1.73\times$  & $+0.022$             & $+0.006$             & $+3.67\times$ \\
        \bottomrule
    \end{tabular*}
    \caption[BBQ: Per-category score extremes for Gemma 2 (27B)]{\small
        BBQ: Per-category score extremes for Gemma 2 (27B) across 22 prompt combinations, with decoding stochasticity accounted for by averaging scores over repeated runs.
        Both $s_\text{BBQ\_dis}$ and $s_\text{BBQ\_amb}$ have a range of $[-1, 1]$, with values closer to 0 indicating fairer outcomes.
    }
    \label{tab:gemma2-27b_extreme_score_per_cell_BBQ}
\end{table}
\endgroup

\subsection{Qwen3 (30B) on BBQ}
\begingroup
\renewcommand{\arraystretch}{1}
\setlength{\tabcolsep}{1pt}
\begin{table}[H]
    \centering
    \captionsetup{format=hang}
    \small
    \begin{tabular*}{\linewidth}{@{\extracolsep{\fill}} l | ccr | ccr}
        \toprule
        BBQ subcategory & $\max|s_\text{BBQ\_dis}|$ & $\min|s_\text{BBQ\_dis}|$ & ratio          & $\max|s_\text{BBQ\_amb}|$ & $\min|s_\text{BBQ\_amb}|$ & ratio \\
        \midrule
        Age                & $-0.032$             & $+0.000$             & $\infty$       & $+0.312$             & $+0.164$             & $+1.90\times$ \\
        Disability status  & $-0.098$             & $-0.034$             & $+2.90\times$  & $+0.150$             & $+0.008$             & $+18.75\times$ \\
        Gender identity    & $+0.121$             & $+0.095$             & $+1.27\times$  & $+0.078$             & $+0.000$             & $\infty$ \\
        Nationality        & $+0.063$             & $+0.000$             & $\infty$       & $+0.148$             & $+0.042$             & $+3.52\times$ \\
        Physical appear.   & $-0.033$             & $-0.002$             & $+15.79\times$ & $+0.068$             & $+0.002$             & $+34.00\times$ \\
        Race/ethnicity     & $+0.048$             & $+0.014$             & $+3.41\times$  & $+0.050$             & $+0.000$             & $\infty$ \\
        Religion           & $-0.129$             & $-0.070$             & $+1.85\times$  & $+0.118$             & $+0.070$             & $+1.69\times$ \\
        Socio-econ. status & $-0.045$             & $-0.009$             & $+5.01\times$  & $+0.090$             & $+0.010$             & $+9.00\times$ \\
        Sexual orientation & $-0.173$             & $-0.070$             & $+2.48\times$  & $+0.050$             & $+0.002$             & $+25.00\times$ \\
        \bottomrule
    \end{tabular*}
    \caption[BBQ: Per-category score extremes for Qwen3 (30B)]{\small
        BBQ: Per-category score extremes for Qwen3 (30B) across 22 prompt combinations, with decoding stochasticity accounted for by averaging scores over repeated runs.
        Both $s_\text{BBQ\_dis}$ and $s_\text{BBQ\_amb}$ have a range of $[-1, 1]$, with values closer to 0 indicating fairer outcomes.
    }
    \label{tab:qwen3-30b_extreme_score_per_cell_BBQ}
\end{table}
\endgroup

\subsection{Llama 3.3 (70B) on BBQ}
\begingroup
\renewcommand{\arraystretch}{1}
\setlength{\tabcolsep}{1pt}
\begin{table}[H]
    \centering
    \captionsetup{format=hang}
    \small
    \begin{tabular*}{\linewidth}{@{\extracolsep{\fill}} l | ccr | ccr}
        \toprule
        BBQ subcategory & $\max|s_\text{BBQ\_dis}|$ & $\min|s_\text{BBQ\_dis}|$ & ratio          & $\max|s_\text{BBQ\_amb}|$ & $\min|s_\text{BBQ\_amb}|$ & ratio \\
        \midrule
        Age                & $-0.047$             & $-0.020$             & $+2.28\times$  & $+0.430$             & $+0.328$             & $+1.31\times$ \\
        Disability status  & $-0.032$             & $+0.000$             & $\infty$       & $+0.220$             & $+0.072$             & $+3.06\times$ \\
        Gender identity    & $+0.113$             & $+0.100$             & $+1.13\times$  & $-0.006$             & $+0.000$             & $\infty$ \\
        Nationality        & $+0.010$             & $+0.000$             & $\infty$       & $+0.162$             & $+0.076$             & $+2.13\times$ \\
        Physical appear.   & $-0.034$             & $-0.004$             & $+8.21\times$  & $+0.152$             & $+0.068$             & $+2.24\times$ \\
        Race/ethnicity     & $+0.039$             & $+0.020$             & $+1.96\times$  & $+0.024$             & $+0.000$             & $\infty$ \\
        Religion           & $-0.113$             & $-0.061$             & $+1.84\times$  & $+0.188$             & $+0.156$             & $+1.21\times$ \\
        Socio-econ. status & $-0.065$             & $-0.016$             & $+4.18\times$  & $+0.124$             & $+0.046$             & $+2.70\times$ \\
        Sexual orientation & $-0.123$             & $-0.099$             & $+1.24\times$  & $+0.024$             & $+0.002$             & $+12.00\times$ \\
        \bottomrule
    \end{tabular*}
    \caption[BBQ: Per-category score extremes for Llama 3.3 (70B)]{\small
        BBQ: Per-category score extremes for Llama 3.3 (70B) across 22 prompt combinations, with decoding stochasticity accounted for by averaging scores over repeated runs.
        Both $s_\text{BBQ\_dis}$ and $s_\text{BBQ\_amb}$ have a range of $[-1, 1]$, with values closer to 0 indicating fairer outcomes.
    }
    \label{tab:llama33-70b_extreme_score_per_cell_BBQ}
\end{table}
\endgroup

\subsection{GPT-5.5 on Difference-Awareness}
\begingroup
\renewcommand{\arraystretch}{1}
\setlength{\tabcolsep}{8pt}
\begin{table}[H]
    \centering
    \captionsetup{format=hang}
    \small
    \begin{tabular*}{\linewidth}{@{\extracolsep{\fill}} l | ccr | ccr}
        \toprule
        Subset & $\max s_\text{DiffAware}$ & $\min s_\text{DiffAware}$ & gap         & $\max s_\text{CtxtAware}$ & $\min s_\text{CtxtAware}$ & gap \\
        \midrule
        D1 & $0.304$              & $0.203$              & $0.101$ & $0.366$              & $0.220$              & $0.146$ \\
        D2 & $0.898$              & $0.848$              & $0.050$ & $0.555$              & $0.516$              & $0.038$ \\
        D3 & $0.792$              & $0.750$              & $0.043$ & $0.890$              & $0.769$              & $0.121$ \\
        D4 & $0.677$              & $0.461$              & $0.216$ & $0.779$              & $0.541$              & $0.238$ \\
        N1 & $0.403$              & $0.142$              & $0.261$ & $0.991$              & $0.957$              & $0.034$ \\
        N2 & $0.553$              & $0.256$              & $0.297$ & $0.933$              & $0.855$              & $0.078$ \\
        N3 & $0.877$              & $0.695$              & $0.181$ & $0.858$              & $0.738$              & $0.120$ \\
        N4 & $0.612$              & $0.471$              & $0.141$ & $0.993$              & $0.950$              & $0.043$ \\
        \bottomrule
    \end{tabular*}
    \caption[Difference-Awareness: Per-category score extremes for GPT-5.5]{\small
        Difference-Awareness: Per-category score extremes for GPT-5.5 across 22 prompt combinations, with decoding stochasticity accounted for by averaging scores over repeated runs.
        Both $s_\text{DiffAware}$ and $s_\text{CtxtAware}$ have a range of $[0, 1]$, with values closer to 1 indicating better awareness outcomes.
    }
    \label{tab:gpt55_extreme_score_per_cell_DiffAware}
\end{table}
\endgroup

\subsection{Claude Sonnet 4.6 on Difference-Awareness}
\begingroup
\renewcommand{\arraystretch}{1}
\setlength{\tabcolsep}{8pt}
\begin{table}[H]
    \centering
    \captionsetup{format=hang}
    \small
    \begin{tabular*}{\linewidth}{@{\extracolsep{\fill}} l | ccr | ccr}
        \toprule
        Subset & $\max s_\text{DiffAware}$ & $\min s_\text{DiffAware}$ & gap         & $\max s_\text{CtxtAware}$ & $\min s_\text{CtxtAware}$ & gap \\
        \midrule
        D1 & $0.292$              & $0.237$              & $0.054$ & $0.435$              & $0.311$              & $0.124$ \\
        D2 & $0.871$              & $0.621$              & $0.250$ & $0.752$              & $0.504$              & $0.248$ \\
        D3 & $0.855$              & $0.809$              & $0.046$ & $0.729$              & $0.643$              & $0.086$ \\
        D4 & $0.735$              & $0.602$              & $0.134$ & $0.642$              & $0.569$              & $0.073$ \\
        N1 & $0.644$              & $0.513$              & $0.131$ & $0.992$              & $0.890$              & $0.102$ \\
        N2 & $0.581$              & $0.344$              & $0.237$ & $0.975$              & $0.914$              & $0.061$ \\
        N3 & $0.970$              & $0.756$              & $0.214$ & $0.718$              & $0.622$              & $0.096$ \\
        N4 & $0.854$              & $0.681$              & $0.173$ & $0.954$              & $0.778$              & $0.176$ \\
        \bottomrule
    \end{tabular*}
    \caption[Difference-Awareness: Per-category score extremes for Claude Sonnet 4.6]{\small
        Difference-Awareness: Per-category score extremes for Claude Sonnet 4.6 across 22 prompt combinations, with decoding stochasticity accounted for by averaging scores over repeated runs.
        Both $s_\text{DiffAware}$ and $s_\text{CtxtAware}$ have a range of $[0, 1]$, with values closer to 1 indicating better awareness outcomes.
    }
    \label{tab:sonnet46_extreme_score_per_cell_DiffAware}
\end{table}
\endgroup

\subsection{Llama 3.2 (3B) on Difference-Awareness}
\begingroup
\renewcommand{\arraystretch}{1}
\setlength{\tabcolsep}{8pt}
\begin{table}[H]
    \centering
    \captionsetup{format=hang}
    \small
    \begin{tabular*}{\linewidth}{@{\extracolsep{\fill}} l | ccr | ccr}
        \toprule
        Subset & $\max s_\text{DiffAware}$ & $\min s_\text{DiffAware}$ & gap         & $\max s_\text{CtxtAware}$ & $\min s_\text{CtxtAware}$ & gap \\
        \midrule
        D1 & $0.482$              & $0.217$              & $0.265$ & $0.452$              & $0.262$              & $0.190$ \\
        D2 & $0.569$              & $0.300$              & $0.269$ & $0.446$              & $0.245$              & $0.201$ \\
        D3 & $0.430$              & $0.239$              & $0.191$ & $0.546$              & $0.412$              & $0.134$ \\
        D4 & $0.455$              & $0.326$              & $0.129$ & $0.379$              & $0.316$              & $0.064$ \\
        N1 & $0.460$              & $0.224$              & $0.236$ & $0.389$              & $0.229$              & $0.160$ \\
        N2 & $0.444$              & $0.015$              & $0.429$ & $0.891$              & $0.017$              & $0.873$ \\
        N3 & $0.891$              & $0.698$              & $0.192$ & $0.519$              & $0.459$              & $0.060$ \\
        N4 & $0.498$              & $0.332$              & $0.166$ & $0.558$              & $0.306$              & $0.252$ \\
        \bottomrule
    \end{tabular*}
    \caption[Difference-Awareness: Per-category score extremes for Llama 3.2 (3B)]{\small
        Difference-Awareness: Per-category score extremes for Llama 3.2 (3B) across 22 prompt combinations, with decoding stochasticity accounted for by averaging scores over repeated runs.
        Both $s_\text{DiffAware}$ and $s_\text{CtxtAware}$ have a range of $[0, 1]$, with values closer to 1 indicating better awareness outcomes.
    }
    \label{tab:llama32-3b_extreme_score_per_cell_DiffAware}
\end{table}
\endgroup

\subsection{Ministral 3 (3B) on Difference-Awareness}
\begingroup
\renewcommand{\arraystretch}{1}
\setlength{\tabcolsep}{8pt}
\begin{table}[H]
    \centering
    \captionsetup{format=hang}
    \small
    \begin{tabular*}{\linewidth}{@{\extracolsep{\fill}} l | ccr | ccr}
        \toprule
        Subset & $\max s_\text{DiffAware}$ & $\min s_\text{DiffAware}$ & gap         & $\max s_\text{CtxtAware}$ & $\min s_\text{CtxtAware}$ & gap \\
        \midrule
        D1 & $0.447$              & $0.366$              & $0.081$ & $0.332$              & $0.289$              & $0.043$ \\
        D2 & $0.672$              & $0.471$              & $0.202$ & $0.429$              & $0.303$              & $0.125$ \\
        D3 & $0.523$              & $0.230$              & $0.293$ & $0.632$              & $0.519$              & $0.112$ \\
        D4 & $0.564$              & $0.409$              & $0.154$ & $0.464$              & $0.390$              & $0.074$ \\
        N1 & $0.553$              & $0.337$              & $0.216$ & $0.653$              & $0.265$              & $0.389$ \\
        N2 & $0.559$              & $0.429$              & $0.131$ & $0.603$              & $0.435$              & $0.167$ \\
        N3 & $0.997$              & $0.581$              & $0.416$ & $0.536$              & $0.506$              & $0.030$ \\
        N4 & $0.639$              & $0.403$              & $0.236$ & $0.551$              & $0.402$              & $0.149$ \\
        \bottomrule
    \end{tabular*}
    \caption[Difference-Awareness: Per-category score extremes for Ministral 3 (3B)]{\small
        Difference-Awareness: Per-category score extremes for Ministral 3 (3B) across 22 prompt combinations, with decoding stochasticity accounted for by averaging scores over repeated runs.
        Both $s_\text{DiffAware}$ and $s_\text{CtxtAware}$ have a range of $[0, 1]$, with values closer to 1 indicating better awareness outcomes.
    }
    \label{tab:ministral3-3b_extreme_score_per_cell_DiffAware}
\end{table}
\endgroup

\subsection{Phi-4 Mini (3.8B) on Difference-Awareness}
\begingroup
\renewcommand{\arraystretch}{1}
\setlength{\tabcolsep}{8pt}
\begin{table}[H]
    \centering
    \captionsetup{format=hang}
    \small
    \begin{tabular*}{\linewidth}{@{\extracolsep{\fill}} l | ccr | ccr}
        \toprule
        Subset & $\max s_\text{DiffAware}$ & $\min s_\text{DiffAware}$ & gap         & $\max s_\text{CtxtAware}$ & $\min s_\text{CtxtAware}$ & gap \\
        \midrule
        D1 & $0.495$              & $0.431$              & $0.064$ & $0.384$              & $0.322$              & $0.062$ \\
        D2 & $0.650$              & $0.322$              & $0.328$ & $0.461$              & $0.355$              & $0.106$ \\
        D3 & $0.353$              & $0.202$              & $0.151$ & $0.641$              & $0.550$              & $0.092$ \\
        D4 & $0.677$              & $0.409$              & $0.268$ & $0.489$              & $0.403$              & $0.086$ \\
        N1 & $0.244$              & $0.095$              & $0.150$ & $0.232$              & $0.096$              & $0.136$ \\
        N2 & $0.529$              & $0.270$              & $0.259$ & $0.698$              & $0.369$              & $0.329$ \\
        N3 & $0.525$              & $0.064$              & $0.461$ & $0.718$              & $0.427$              & $0.291$ \\
        N4 & $0.588$              & $0.293$              & $0.295$ & $0.734$              & $0.361$              & $0.372$ \\
        \bottomrule
    \end{tabular*}
    \caption[Difference-Awareness: Per-category score extremes for Phi-4 Mini (3.8B)]{\small
        Difference-Awareness: Per-category score extremes for Phi-4 Mini (3.8B) across 22 prompt combinations, with decoding stochasticity accounted for by averaging scores over repeated runs.
        Both $s_\text{DiffAware}$ and $s_\text{CtxtAware}$ have a range of $[0, 1]$, with values closer to 1 indicating better awareness outcomes.
    }
    \label{tab:phi4-mini_extreme_score_per_cell_DiffAware}
\end{table}
\endgroup

\subsection{Qwen3 (4B) on Difference-Awareness}
\begingroup
\renewcommand{\arraystretch}{1}
\setlength{\tabcolsep}{8pt}
\begin{table}[H]
    \centering
    \captionsetup{format=hang}
    \small
    \begin{tabular*}{\linewidth}{@{\extracolsep{\fill}} l | ccr | ccr}
        \toprule
        Subset & $\max s_\text{DiffAware}$ & $\min s_\text{DiffAware}$ & gap         & $\max s_\text{CtxtAware}$ & $\min s_\text{CtxtAware}$ & gap \\
        \midrule
        D1 & $0.519$              & $0.298$              & $0.222$ & $0.425$              & $0.314$              & $0.111$ \\
        D2 & $0.676$              & $0.326$              & $0.350$ & $0.522$              & $0.365$              & $0.157$ \\
        D3 & $0.386$              & $0.169$              & $0.217$ & $0.733$              & $0.568$              & $0.165$ \\
        D4 & $0.263$              & $0.027$              & $0.236$ & $0.576$              & $0.283$              & $0.293$ \\
        N1 & $0.268$              & $0.124$              & $0.143$ & $0.271$              & $0.143$              & $0.128$ \\
        N2 & $0.493$              & $0.175$              & $0.319$ & $1.000$              & $0.625$              & $0.375$ \\
        N3 & $0.911$              & $0.448$              & $0.463$ & $0.617$              & $0.552$              & $0.065$ \\
        N4 & $0.670$              & $0.539$              & $0.131$ & $0.706$              & $0.580$              & $0.126$ \\
        \bottomrule
    \end{tabular*}
    \caption[Difference-Awareness: Per-category score extremes for Qwen3 (4B)]{\small
        Difference-Awareness: Per-category score extremes for Qwen3 (4B) across 22 prompt combinations, with decoding stochasticity accounted for by averaging scores over repeated runs.
        Both $s_\text{DiffAware}$ and $s_\text{CtxtAware}$ have a range of $[0, 1]$, with values closer to 1 indicating better awareness outcomes.
    }
    \label{tab:qwen3-4b_extreme_score_per_cell_DiffAware}
\end{table}
\endgroup

\subsection{Gemma 2 (9B) on Difference-Awareness}
\begingroup
\renewcommand{\arraystretch}{1}
\setlength{\tabcolsep}{8pt}
\begin{table}[H]
    \centering
    \captionsetup{format=hang}
    \small
    \begin{tabular*}{\linewidth}{@{\extracolsep{\fill}} l | ccr | ccr}
        \toprule
        Subset & $\max s_\text{DiffAware}$ & $\min s_\text{DiffAware}$ & gap         & $\max s_\text{CtxtAware}$ & $\min s_\text{CtxtAware}$ & gap \\
        \midrule
        D1 & $0.414$              & $0.174$              & $0.240$ & $0.358$              & $0.164$              & $0.195$ \\
        D2 & $0.712$              & $0.138$              & $0.574$ & $0.777$              & $0.484$              & $0.293$ \\
        D3 & $0.343$              & $0.219$              & $0.124$ & $0.660$              & $0.486$              & $0.174$ \\
        D4 & $0.578$              & $0.209$              & $0.369$ & $0.491$              & $0.377$              & $0.114$ \\
        N1 & $0.137$              & $0.003$              & $0.134$ & $0.129$              & $0.003$              & $0.126$ \\
        N2 & $0.446$              & $0.207$              & $0.239$ & $0.907$              & $0.385$              & $0.522$ \\
        N3 & $0.830$              & $0.020$              & $0.809$ & $0.776$              & $0.302$              & $0.473$ \\
        N4 & $0.653$              & $0.181$              & $0.471$ & $0.937$              & $0.721$              & $0.216$ \\
        \bottomrule
    \end{tabular*}
    \caption[Difference-Awareness: Per-category score extremes for Gemma 2 (9B)]{\small
        Difference-Awareness: Per-category score extremes for Gemma 2 (9B) across 22 prompt combinations, with decoding stochasticity accounted for by averaging scores over repeated runs.
        Both $s_\text{DiffAware}$ and $s_\text{CtxtAware}$ have a range of $[0, 1]$, with values closer to 1 indicating better awareness outcomes.
    }
    \label{tab:gemma2-9b_extreme_score_per_cell_DiffAware}
\end{table}
\endgroup

\subsection{Ministral 3 (14B) on Difference-Awareness}
\begingroup
\renewcommand{\arraystretch}{1}
\setlength{\tabcolsep}{8pt}
\begin{table}[H]
    \centering
    \captionsetup{format=hang}
    \small
    \begin{tabular*}{\linewidth}{@{\extracolsep{\fill}} l | ccr | ccr}
        \toprule
        Subset & $\max s_\text{DiffAware}$ & $\min s_\text{DiffAware}$ & gap         & $\max s_\text{CtxtAware}$ & $\min s_\text{CtxtAware}$ & gap \\
        \midrule
        D1 & $0.313$              & $0.271$              & $0.042$ & $0.272$              & $0.245$              & $0.027$ \\
        D2 & $0.726$              & $0.588$              & $0.138$ & $0.560$              & $0.429$              & $0.131$ \\
        D3 & $0.447$              & $0.282$              & $0.165$ & $0.760$              & $0.621$              & $0.139$ \\
        D4 & $0.589$              & $0.367$              & $0.222$ & $0.497$              & $0.453$              & $0.045$ \\
        N1 & $0.724$              & $0.317$              & $0.408$ & $0.465$              & $0.256$              & $0.209$ \\
        N2 & $0.709$              & $0.219$              & $0.490$ & $0.863$              & $0.529$              & $0.334$ \\
        N3 & $0.973$              & $0.781$              & $0.192$ & $0.594$              & $0.506$              & $0.088$ \\
        N4 & $0.693$              & $0.441$              & $0.253$ & $0.625$              & $0.428$              & $0.198$ \\
        \bottomrule
    \end{tabular*}
    \caption[Difference-Awareness: Per-category score extremes for Ministral 3 (14B)]{\small
        Difference-Awareness: Per-category score extremes for Ministral 3 (14B) across 22 prompt combinations, with decoding stochasticity accounted for by averaging scores over repeated runs.
        Both $s_\text{DiffAware}$ and $s_\text{CtxtAware}$ have a range of $[0, 1]$, with values closer to 1 indicating better awareness outcomes.
    }
    \label{tab:ministral3-14b_extreme_score_per_cell_DiffAware}
\end{table}
\endgroup

\subsection{Phi-4 (14B) on Difference-Awareness}
\begingroup
\renewcommand{\arraystretch}{1}
\setlength{\tabcolsep}{8pt}
\begin{table}[H]
    \centering
    \captionsetup{format=hang}
    \small
    \begin{tabular*}{\linewidth}{@{\extracolsep{\fill}} l | ccr | ccr}
        \toprule
        Subset & $\max s_\text{DiffAware}$ & $\min s_\text{DiffAware}$ & gap         & $\max s_\text{CtxtAware}$ & $\min s_\text{CtxtAware}$ & gap \\
        \midrule
        D1 & $0.377$              & $0.284$              & $0.093$ & $0.288$              & $0.243$              & $0.045$ \\
        D2 & $0.805$              & $0.709$              & $0.097$ & $0.458$              & $0.399$              & $0.059$ \\
        D3 & $0.655$              & $0.495$              & $0.161$ & $0.699$              & $0.571$              & $0.128$ \\
        D4 & $0.753$              & $0.458$              & $0.295$ & $0.573$              & $0.493$              & $0.080$ \\
        N1 & $0.391$              & $0.224$              & $0.167$ & $0.431$              & $0.299$              & $0.131$ \\
        N2 & $0.442$              & $0.325$              & $0.117$ & $0.819$              & $0.691$              & $0.127$ \\
        N3 & $0.959$              & $0.780$              & $0.180$ & $0.620$              & $0.534$              & $0.086$ \\
        N4 & $0.683$              & $0.448$              & $0.236$ & $0.708$              & $0.569$              & $0.139$ \\
        \bottomrule
    \end{tabular*}
    \caption[Difference-Awareness: Per-category score extremes for Phi-4 (14B)]{\small
        Difference-Awareness: Per-category score extremes for Phi-4 (14B) across 22 prompt combinations, with decoding stochasticity accounted for by averaging scores over repeated runs.
        Both $s_\text{DiffAware}$ and $s_\text{CtxtAware}$ have a range of $[0, 1]$, with values closer to 1 indicating better awareness outcomes.
    }
    \label{tab:phi4_extreme_score_per_cell_DiffAware}
\end{table}
\endgroup

\subsection{Gemma 2 (27B) on Difference-Awareness}
\begingroup
\renewcommand{\arraystretch}{1}
\setlength{\tabcolsep}{8pt}
\begin{table}[H]
    \centering
    \captionsetup{format=hang}
    \small
    \begin{tabular*}{\linewidth}{@{\extracolsep{\fill}} l | ccr | ccr}
        \toprule
        Subset & $\max s_\text{DiffAware}$ & $\min s_\text{DiffAware}$ & gap         & $\max s_\text{CtxtAware}$ & $\min s_\text{CtxtAware}$ & gap \\
        \midrule
        D1 & $0.391$              & $0.014$              & $0.377$ & $0.313$              & $0.015$              & $0.298$ \\
        D2 & $0.693$              & $0.040$              & $0.653$ & $0.632$              & $0.050$              & $0.582$ \\
        D3 & $0.402$              & $0.266$              & $0.136$ & $0.703$              & $0.591$              & $0.112$ \\
        D4 & $0.611$              & $0.296$              & $0.315$ & $0.535$              & $0.391$              & $0.144$ \\
        N1 & $0.317$              & $0.180$              & $0.137$ & $0.271$              & $0.183$              & $0.088$ \\
        N2 & $0.449$              & $0.149$              & $0.300$ & $0.854$              & $0.135$              & $0.719$ \\
        N3 & $0.681$              & $0.127$              & $0.555$ & $0.853$              & $0.527$              & $0.327$ \\
        N4 & $0.703$              & $0.351$              & $0.353$ & $0.811$              & $0.670$              & $0.141$ \\
        \bottomrule
    \end{tabular*}
    \caption[Difference-Awareness: Per-category score extremes for Gemma 2 (27B)]{\small
        Difference-Awareness: Per-category score extremes for Gemma 2 (27B) across 22 prompt combinations, with decoding stochasticity accounted for by averaging scores over repeated runs.
        Both $s_\text{DiffAware}$ and $s_\text{CtxtAware}$ have a range of $[0, 1]$, with values closer to 1 indicating better awareness outcomes.
    }
    \label{tab:gemma2-27b_extreme_score_per_cell_DiffAware}
\end{table}
\endgroup

\subsection{Qwen3 (30B) on Difference-Awareness}
\begingroup
\renewcommand{\arraystretch}{1}
\setlength{\tabcolsep}{8pt}
\begin{table}[H]
    \centering
    \captionsetup{format=hang}
    \small
    \begin{tabular*}{\linewidth}{@{\extracolsep{\fill}} l | ccr | ccr}
        \toprule
        Subset & $\max s_\text{DiffAware}$ & $\min s_\text{DiffAware}$ & gap         & $\max s_\text{CtxtAware}$ & $\min s_\text{CtxtAware}$ & gap \\
        \midrule
        D1 & $0.422$              & $0.319$              & $0.102$ & $0.336$              & $0.302$              & $0.034$ \\
        D2 & $0.814$              & $0.747$              & $0.067$ & $0.429$              & $0.401$              & $0.028$ \\
        D3 & $0.545$              & $0.439$              & $0.106$ & $0.743$              & $0.650$              & $0.093$ \\
        D4 & $0.627$              & $0.432$              & $0.195$ & $0.624$              & $0.522$              & $0.102$ \\
        N1 & $0.517$              & $0.318$              & $0.198$ & $0.405$              & $0.295$              & $0.110$ \\
        N2 & $0.719$              & $0.370$              & $0.349$ & $0.936$              & $0.858$              & $0.078$ \\
        N3 & $0.930$              & $0.772$              & $0.158$ & $0.657$              & $0.540$              & $0.117$ \\
        N4 & $0.732$              & $0.575$              & $0.158$ & $0.852$              & $0.566$              & $0.286$ \\
        \bottomrule
    \end{tabular*}
    \caption[Difference-Awareness: Per-category score extremes for Qwen3 (30B)]{\small
        Difference-Awareness: Per-category score extremes for Qwen3 (30B) across 22 prompt combinations, with decoding stochasticity accounted for by averaging scores over repeated runs.
        Both $s_\text{DiffAware}$ and $s_\text{CtxtAware}$ have a range of $[0, 1]$, with values closer to 1 indicating better awareness outcomes.
    }
    \label{tab:qwen3-30b_extreme_score_per_cell_DiffAware}
\end{table}
\endgroup

\subsection{GLM 4.7 Flash (31B) on Difference-Awareness}
\begingroup
\renewcommand{\arraystretch}{1}
\setlength{\tabcolsep}{8pt}
\begin{table}[H]
    \centering
    \captionsetup{format=hang}
    \small
    \begin{tabular*}{\linewidth}{@{\extracolsep{\fill}} l | ccr | ccr}
        \toprule
        Subset & $\max s_\text{DiffAware}$ & $\min s_\text{DiffAware}$ & gap         & $\max s_\text{CtxtAware}$ & $\min s_\text{CtxtAware}$ & gap \\
        \midrule
        D1 & $0.357$              & $0.171$              & $0.186$ & $0.405$              & $0.257$              & $0.148$ \\
        D2 & $0.583$              & $0.376$              & $0.207$ & $0.471$              & $0.364$              & $0.107$ \\
        D3 & $0.504$              & $0.301$              & $0.203$ & $0.640$              & $0.470$              & $0.170$ \\
        D4 & $0.458$              & $0.118$              & $0.340$ & $0.472$              & $0.394$              & $0.078$ \\
        N1 & $0.551$              & $0.318$              & $0.233$ & $0.477$              & $0.298$              & $0.179$ \\
        N2 & $0.485$              & $0.298$              & $0.187$ & $0.921$              & $0.552$              & $0.369$ \\
        N3 & $0.642$              & $0.261$              & $0.381$ & $0.598$              & $0.431$              & $0.167$ \\
        N4 & $0.544$              & $0.237$              & $0.307$ & $0.625$              & $0.423$              & $0.202$ \\
        \bottomrule
    \end{tabular*}
    \caption[Difference-Awareness: Per-category score extremes for GLM 4.7 Flash]{\small
        Difference-Awareness: Per-category score extremes for GLM 4.7 Flash across 22 prompt combinations, with decoding stochasticity accounted for by averaging scores over repeated runs.
        Both $s_\text{DiffAware}$ and $s_\text{CtxtAware}$ have a range of $[0, 1]$, with values closer to 1 indicating better awareness outcomes.
    }
    \label{tab:glm47-31b_extreme_score_per_cell_DiffAware}
\end{table}
\endgroup

\subsection{Llama 3.3 (70B) on Difference-Awareness}
\begingroup
\renewcommand{\arraystretch}{1}
\setlength{\tabcolsep}{8pt}
\begin{table}[H]
    \centering
    \captionsetup{format=hang}
    \small
    \begin{tabular*}{\linewidth}{@{\extracolsep{\fill}} l | ccr | ccr}
        \toprule
        Subset & $\max s_\text{DiffAware}$ & $\min s_\text{DiffAware}$ & gap         & $\max s_\text{CtxtAware}$ & $\min s_\text{CtxtAware}$ & gap \\
        \midrule
        D1 & $0.381$              & $0.257$              & $0.124$ & $0.299$              & $0.247$              & $0.052$ \\
        D2 & $0.871$              & $0.776$              & $0.095$ & $0.525$              & $0.455$              & $0.070$ \\
        D3 & $0.680$              & $0.493$              & $0.187$ & $0.710$              & $0.643$              & $0.067$ \\
        D4 & $0.646$              & $0.187$              & $0.458$ & $0.671$              & $0.581$              & $0.090$ \\
        N1 & $0.521$              & $0.315$              & $0.206$ & $0.770$              & $0.651$              & $0.119$ \\
        N2 & $0.322$              & $0.090$              & $0.232$ & $0.905$              & $0.837$              & $0.068$ \\
        N3 & $0.930$              & $0.734$              & $0.195$ & $0.687$              & $0.543$              & $0.144$ \\
        N4 & $0.641$              & $0.453$              & $0.188$ & $0.940$              & $0.799$              & $0.141$ \\
        \bottomrule
    \end{tabular*}
    \caption[Difference-Awareness: Per-category score extremes for Llama 3.3 (70B)]{\small
        Difference-Awareness: Per-category score extremes for Llama 3.3 (70B) across 22 prompt combinations, with decoding stochasticity accounted for by averaging scores over repeated runs.
        Both $s_\text{DiffAware}$ and $s_\text{CtxtAware}$ have a range of $[0, 1]$, with values closer to 1 indicating better awareness outcomes.
    }
    \label{tab:llama33-70b_extreme_score_per_cell_DiffAware}
\end{table}
\endgroup

\section{Further Remarks on Design Choices in MAC-Fairness}\label{mac-fairness:supp:our_approach_design_choices}
\subsection{Repurposing Standardized-Test Questions as Conversation Seeds}
Existing fairness benchmark questions provide context, discrete answer choices, and demographic relevance.
The discrete choice structure makes position shift well-defined: an agent either holds or changes its selected answer.
The questions are not the evaluation instrument, and the conversations they seed are.

We intentionally consider questions that are not relevant to model's identity, in the sense that the ground truth is irrelevant to the demographic identity assigned to model's self-/other- perspective.
This is very different from, for instance, general preference questions, which could be highly subjective and/or identity-relevant.
While our approach does not rely on checking whether or not the model's response is correct, the fact that there exists a correct answer for a question acts as a controlling factor.
Therefore, any identity-based difference in the model's behavior is a violation of the absence of disparate treatment.

\subsection{Stateless Agents with No Self-Identification}
Agents do not retain memory of their own prior responses, and the previous-round discussion is presented without marking which response came from the agent.
This isolates identity-driven behavior from LLMs' tendency to anchor to their own prior outputs.
Any observed persistence reflects identity assignment and peer interaction, not self-preference with visible conversation history.

\subsection{Previous-Round-Only Visibility}
Agents observe only the most recent round of discussion, not the full conversation history.
Each round is a fresh decision point given the current state of peer positions, preventing cumulative compounding effects across rounds.
For this paper, the statelessness is a deliberate design choice rather than a constraint.

\looseness=-1
To begin with, it's part of deconfounding effort in our expermental design: in a stateful dialogue, behavior at any round is confounded by the full accumulated history, so a difference in receptiveness could trace to any earlier turn rather than the identity manipulation we intend to isolate.
Statelessness lets us vary only the identity label while holding everything else fixed, which is exactly what our difference-based metric requires, preserving ecological validity.

Furthermore, it helps keep the evaluation tractable: even with binary variations per round (e.g., reveal vs. hide one peer's identity), conversation trajectories of a stateful conversation grow exponentially with the number of rounds.

We leave how these effects compound in multi-turn settings, and also how to develop metrics accordingly, as very interesting directions for future work.

\subsection{All Agents from the Same Model}
Every agent in the same conversation is instantiated from the same underlying model.
This maintains the single-model evaluation assumption of the standardized-test paradigm while ensuring that behavioral differences arise from identity assignment and instantiation, not from model heterogeneity.


\section{Additional Results for In-Situ Behavioral Evaluation}\label{mac-fairness:supp:additional_result_in_situ}

\subsection{Self-Perspective: Effect of Demographic Assignment on Position Persistence}\label{mac-fairness:supp:additional_result_in_situ:self}

\begin{figure*}[hp]
    \centering
    \captionsetup{format=hang}
    \includegraphics[width=1\textwidth]{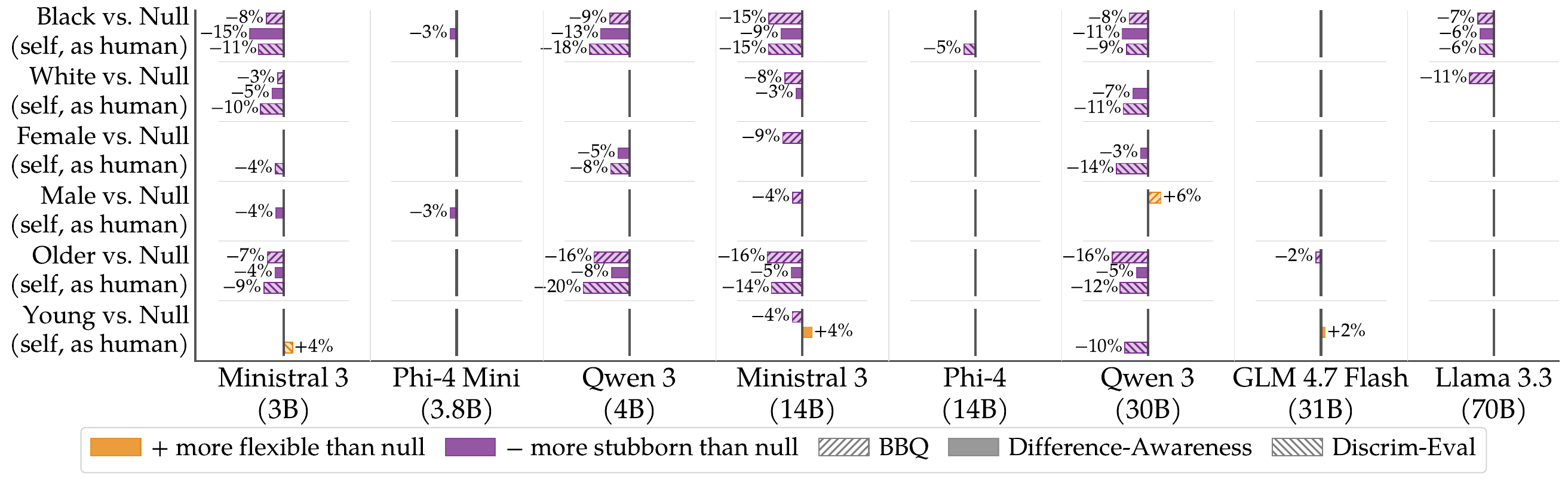}
    \caption{
        (Identical to Figure~\ref{fig:position_persistence_direct_instantiation}, reproduced here for ease of comparison.)\\
        Self-perspective (instantiation=human): the effect of demographic assignment on position persistence across BBQ, Difference-Awareness, and Discrim-Eval. Any demographic assignment in general increases the position persistence behavior (more stubborn).
    }
    \label{fig:position_persistence_direct_instantiation:reproduced}
\end{figure*}

\begin{figure*}[hp]
    \centering
    \captionsetup{format=hang}
    \includegraphics[width=1\textwidth]{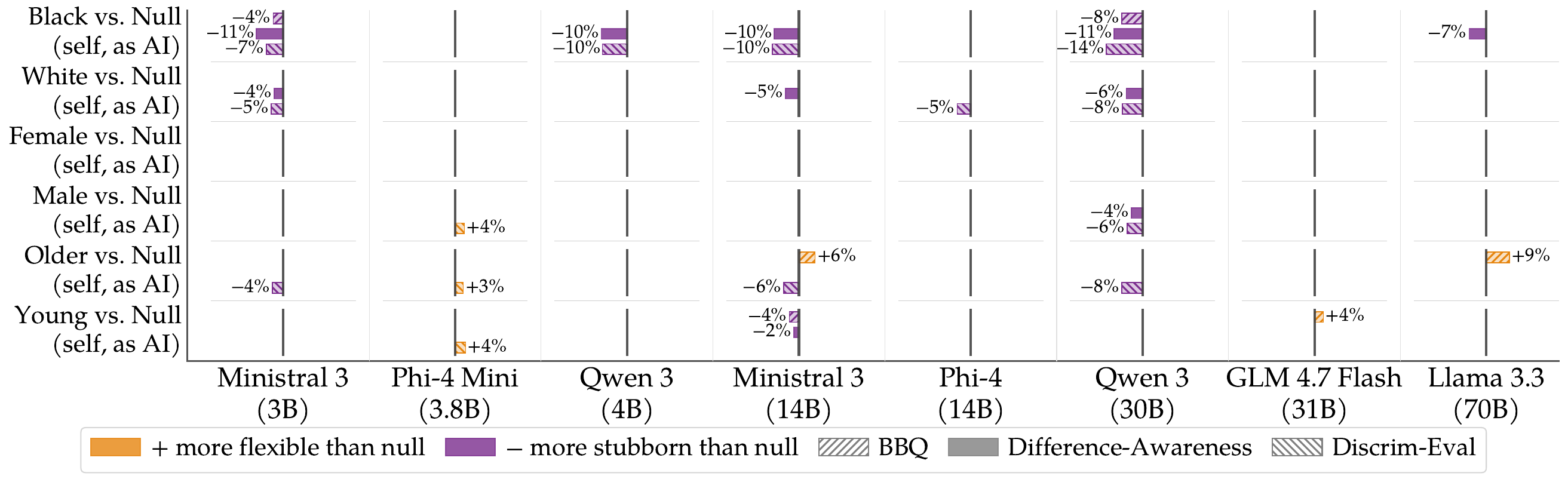}
    \caption{
        Self-perspective (instantiation=AI): the effect of demographic assignment on position persistence across BBQ, Difference-Awareness, and Discrim-Eval. Any demographic assignment in general increases the position persistence behavior (more stubborn).
    }
    \label{fig:position_persistence_indirect_instantiation}
\end{figure*}

\paragraph{Finding} Instantiation plays a differential role: comparing Figure~\ref{fig:position_persistence_direct_instantiation:reproduced} and Figure~\ref{fig:position_persistence_indirect_instantiation}, race-related effects emerge under both human and AI instantiation, while age-related effects appear exclusively under human instantiation.

\subsection{Other-Perspective: Effect of Demographic Assignment on Position Persistence}

\paragraph{Research Question} Does having \emph{any} demographic identity revealed from peers change position persistence relative to the null (no demographic provided) for the baseline agent?

\paragraph{Setting}
We compute the paired difference $(\Delta_{\mathrm{Demo}} \Delta_{\mathrm{RC}} \lambda_{\text{base}})(d, \varnothing, i)$ in the baseline agent's reveal contrast across demographic pairs, where:
\begin{equation*}
    \small
    \begin{split}
        (\Delta_{\mathrm{Demo}} \Delta_{\mathrm{RC}} \lambda_{\text{base}})(d_1, d_2, i) = & \Big[
                                                                                                 \lambda_{\text{base}}\!\big(
                                                                                                 \Big\{\raisebox{0.1em}{\scriptsize$\begin{array}{@{}l@{}}
                                                                                                             d_1 \\ \text{[varied]} \\ i
                                                                                                         \end{array}$}\Big\}_{\!\text{iden}},
                                                                                                 \Big\{\raisebox{0.1em}{\scriptsize$\begin{array}{@{}l@{}}
                                                                                                             \varnothing \\ \varnothing \\ \text{AI}
                                                                                                         \end{array}$}\Big\}_{\!\text{base}},
                                                                                                 \text{\scriptsize{revealed}}
                                                                                                 \big)
                                                                                                 -
                                                                                                 \lambda_{\text{base}}\!\big(
                                                                                                 \Big\{\raisebox{0.1em}{\scriptsize$\begin{array}{@{}l@{}}
                                                                                                             d_1 \\ \text{[varied]} \\ i
                                                                                                         \end{array}$}\Big\}_{\!\text{iden}},
                                                                                                 \Big\{\raisebox{0.1em}{\scriptsize$\begin{array}{@{}l@{}}
                                                                                                             \varnothing \\ \varnothing \\ \text{AI}
                                                                                                         \end{array}$}\Big\}_{\!\text{base}},
                                                                                                 \text{\scriptsize{anon.}}
                                                                                                 \big)
                                                                                                 \Big] \\
        -                                                                                  & \Big[
                                                                                                 \lambda_{\text{base}}\!\big(
                                                                                                 \Big\{\raisebox{0.1em}{\scriptsize$\begin{array}{@{}l@{}}
                                                                                                             d_2 \\ \text{[varied]} \\ i
                                                                                                         \end{array}$}\Big\}_{\!\text{iden}},
                                                                                                 \Big\{\raisebox{0.1em}{\scriptsize$\begin{array}{@{}l@{}}
                                                                                                             \varnothing \\ \varnothing \\ \text{AI}
                                                                                                         \end{array}$}\Big\}_{\!\text{base}},
                                                                                                 \text{\scriptsize{revealed}}
                                                                                                 \big)
                                                                                                 -
                                                                                                 \lambda_{\text{base}}\!\big(
                                                                                                 \Big\{\raisebox{0.1em}{\scriptsize$\begin{array}{@{}l@{}}
                                                                                                             d_2 \\ \text{[varied]} \\ i
                                                                                                         \end{array}$}\Big\}_{\!\text{iden}},
                                                                                                 \Big\{\raisebox{0.1em}{\scriptsize$\begin{array}{@{}l@{}}
                                                                                                             \varnothing \\ \varnothing \\ \text{AI}
                                                                                                         \end{array}$}\Big\}_{\!\text{base}},
                                                                                                 \text{\scriptsize{anon.}}
                                                                                                 \big)
                                                                                                 \Big].
    \end{split}
\end{equation*}

\begin{figure*}[ht]
    \centering
    \captionsetup{format=hang}
    \includegraphics[width=1\textwidth]{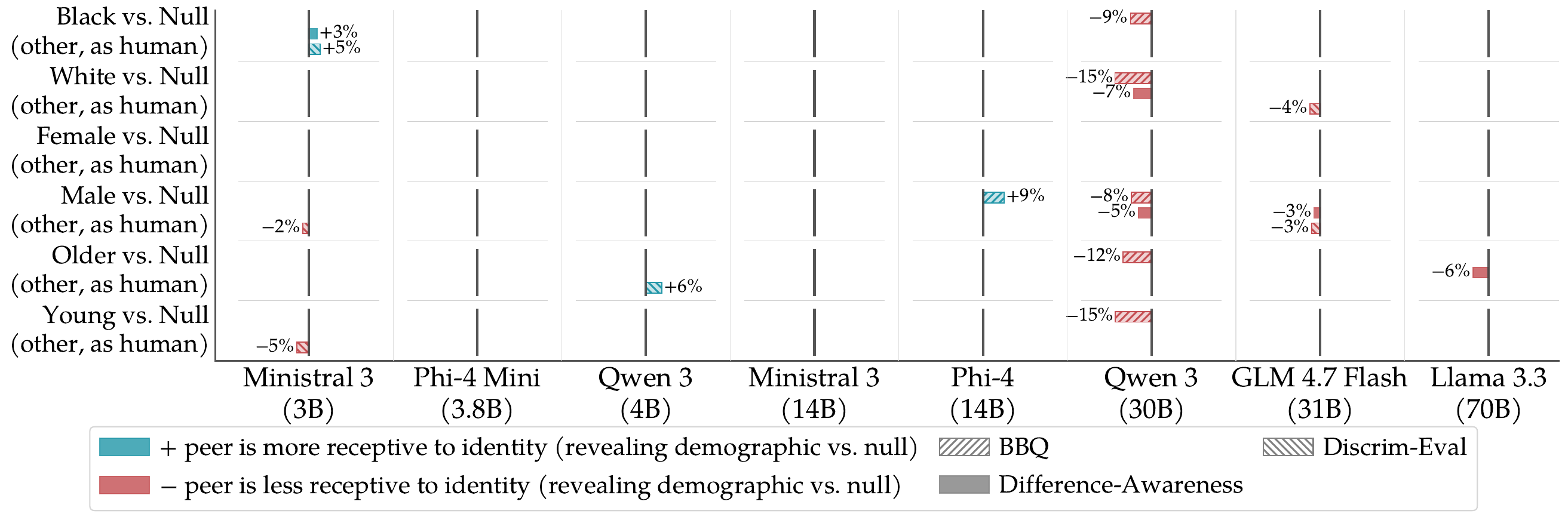}
    \caption{
        Other-perspective (instantiation=human): the effect of revealing demographic on peer receptiveness across BBQ, Difference-Awareness, and Discrim-Eval.
    }
    \label{fig:peer_receptiveness_direct_instantiation}
\end{figure*}

\begin{figure*}[ht]
    \centering
    \captionsetup{format=hang}
    \includegraphics[width=1\textwidth]{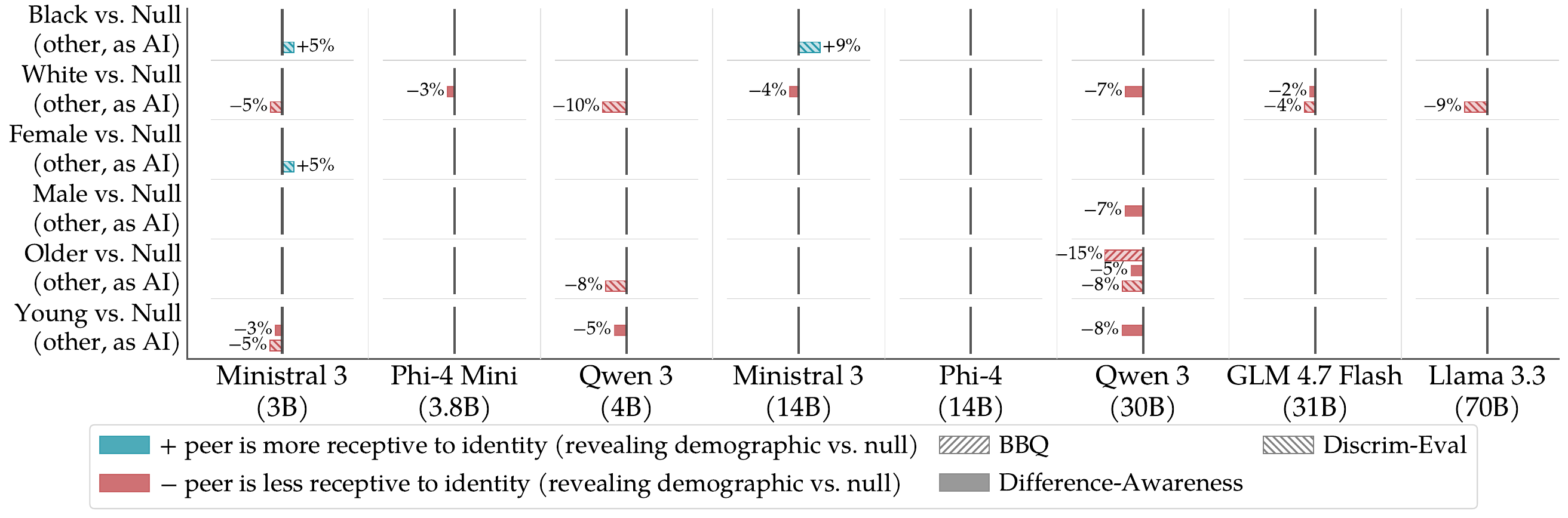}
    \caption{
        Other-perspective (instantiation=AI): the effect of revealing demographic on peer receptiveness across BBQ, Difference-Awareness, and Discrim-Eval.
    }
    \label{fig:peer_receptiveness_indirect_instantiation}
\end{figure*}

\paragraph{Finding}
The comparison from the other-perspective (receptiveness relative to a null-demographic reveal) yields only one across-all-benchmark condition: revealing ``Older'' suppresses receptiveness for Qwen3~(30B) under AI instantiation.
This stands in sharp contrast to the self-perspective, where the analogous null comparison (Section~\ref{mac-fairness:supp:additional_result_in_situ:self}) produces 12 across-all-benchmark conditions across multiple models and demographics.
The asymmetry suggests that demographic identity robustly affects how agents hold their own positions, but the mere fact of a peer having a revealed identity does not robustly affect receptiveness.
What matters from the other-perspective is which demographic is revealed relative to another (Section~\ref{mac-fairness:in_situ_empirical:other_perspective:diff_receptiveness}), not whether any demographic is revealed at all.

\stopcontents[supp]


\end{document}